\documentclass{article}

\usepackage[final]{neurips_2023}

\usepackage{hyperref}

\usepackage[utf8]{inputenc} 
\usepackage[T1]{fontenc}    
\usepackage{hyperref}       
\usepackage{url}            
\usepackage{booktabs}       
\usepackage{amsfonts}       
\usepackage{nicefrac}       
\usepackage{microtype}      
\usepackage{xcolor}         

\usepackage{enumitem}
\usepackage{booktabs}  
\usepackage{graphicx}
\usepackage{caption}
\usepackage{subcaption}
\usepackage{wrapfig}
\usepackage{amssymb}
\usepackage{amsmath}
\usepackage{makecell}

\usepackage{multirow}
\usepackage{placeins}
\usepackage[normalem]{ulem}

\usepackage{wrapfig}
\usepackage{bbm}
\usepackage{amsthm}

\usepackage{amsmath,amsfonts,bm}

\def\rrrl{{\large[\normalsize}}
\def\rrrr{{\large]\normalsize}}

\def\eqref#1{equation~\ref{#1}}

\def\1{\bm{1}}

\def\appnabla{{\widehat{\nabla}}}

\newtheorem{innercustomthm}{Theorem}
\newenvironment{customthm}[1]
  {\renewcommand\theinnercustomthm{#1}\innercustomthm}
  {\endinnercustomthm}
  
\newtheorem{innercustomrmk}{Remark}
\newenvironment{customrmk}[1]
  {\renewcommand\theinnercustomrmk{#1}\innercustomrmk}
  {\endinnercustomrmk}
  
\def\vtheta{{\bm{\theta}}}
\def\vpi{{\bm{\pi}}}
\def\vphi{{\bm{\phi}}}

\def\vc{{\bm{c}}}

\def\mD{{\bm{D}}}

\def\mG{{\bm{G}}}

\def\mI{{\bm{I}}}

\def\mX{{\bm{X}}}

\def\mZero{{\bm{0}}}

\DeclareMathAlphabet{\mathsfit}{\encodingdefault}{\sfdefault}{m}{sl}
\SetMathAlphabet{\mathsfit}{bold}{\encodingdefault}{\sfdefault}{bx}{n}
\newcommand{\tens}[1]{\bm{\mathsfit{#1}}}

\def\tI{{\tens{I}}}

\usepackage[linesnumbered,ruled,vlined]{algorithm2e}

\newtheorem{definition}{Definition}[section]

\newtheorem{theorem}{Theorem}[section]
\newtheorem{remark}{Remark}[section]

\newcommand{\ours}{ReinMax }

\newcommand{\smallsection}[1]{\textbf{#1.~~~~}}

\newcommand{\noise}[1]{implicit label noise\xspace}
\newcommand{\Noise}[1]{Implicit label noise\xspace}

\title{Bridging Discrete and Backpropagation: \\ Straight-Through and Beyond}

\author{%
  Liyuan Liu \; Chengyu Dong \; Xiaodong Liu \; Bin Yu \; Jianfeng Gao  \\
  Microsoft Research \\
  {\small \texttt{\{lucliu, v-chedong, xiaodl, v-ybi, jfgao\}@microsoft.com} }\\
}

\begin{document}

\maketitle

\begin{abstract}
Backpropagation, the cornerstone of deep learning, is limited to computing gradients for continuous variables.
This limitation poses challenges for problems involving discrete latent variables.
To address this issue, we propose a novel approach to approximate the gradient of parameters involved in generating discrete latent variables.  
First, we examine the widely used Straight-Through (ST) heuristic and demonstrate that it works as a first-order approximation of the gradient.
Guided by our findings, we propose ReinMax, which achieves second-order accuracy by integrating Heun's method, a second-order numerical method for solving ODEs.
\ours does not require Hessian or other second-order derivatives, thus having negligible computation overheads.
Extensive experimental results on various tasks demonstrate the superiority of \ours over the state of the art.

\end{abstract}

\section{Introduction}

There has been a persistent pursuit to build neural network models with discrete or sparse variables~\citep{Neal1992ConnectionistLO}. 
However, backpropagation~\citep{Rumelhari2004LearningRB}, the cornerstone of deep learning, is restricted to computing gradients for continuous variables.
Correspondingly, many attempts have been made to 
approximate the gradient of parameters that are used to generate discrete variables, and most of them are based on the Straight-Through (ST) technique~\citep{Bengio2013EstimatingOP}.

The development of ST is based on the simple intuition that non-differentiable functions (e.g., sampling of discrete latent variables) can be approximated with the identity function in the back-propagation~\citep{rosenblatt1957perceptron,Bengio2013EstimatingOP}. 
Due to the lack of theoretical underpinnings, there is neither guarantee that ST can be viewed as an approximation of the gradient, nor guidance on hyper-parameter configurations or future algorithm development. 
Thus, researchers have to develop different ST variants for different applications in a trial-and-error manner, which is laborious and time-consuming~\citep {Oord2017NeuralDR,liu2018darts,Fedus2021SwitchTS}.
To address these limitations, we aim to explore \emph{how ST approximates the gradient and how it can be improved.} 

\begin{figure}[h!]
    \centering 
    \vspace{-2.5mm}
    \begin{subfigure}[t]{0.325\linewidth}
        \centering
        \includegraphics[width=1.0\textwidth]{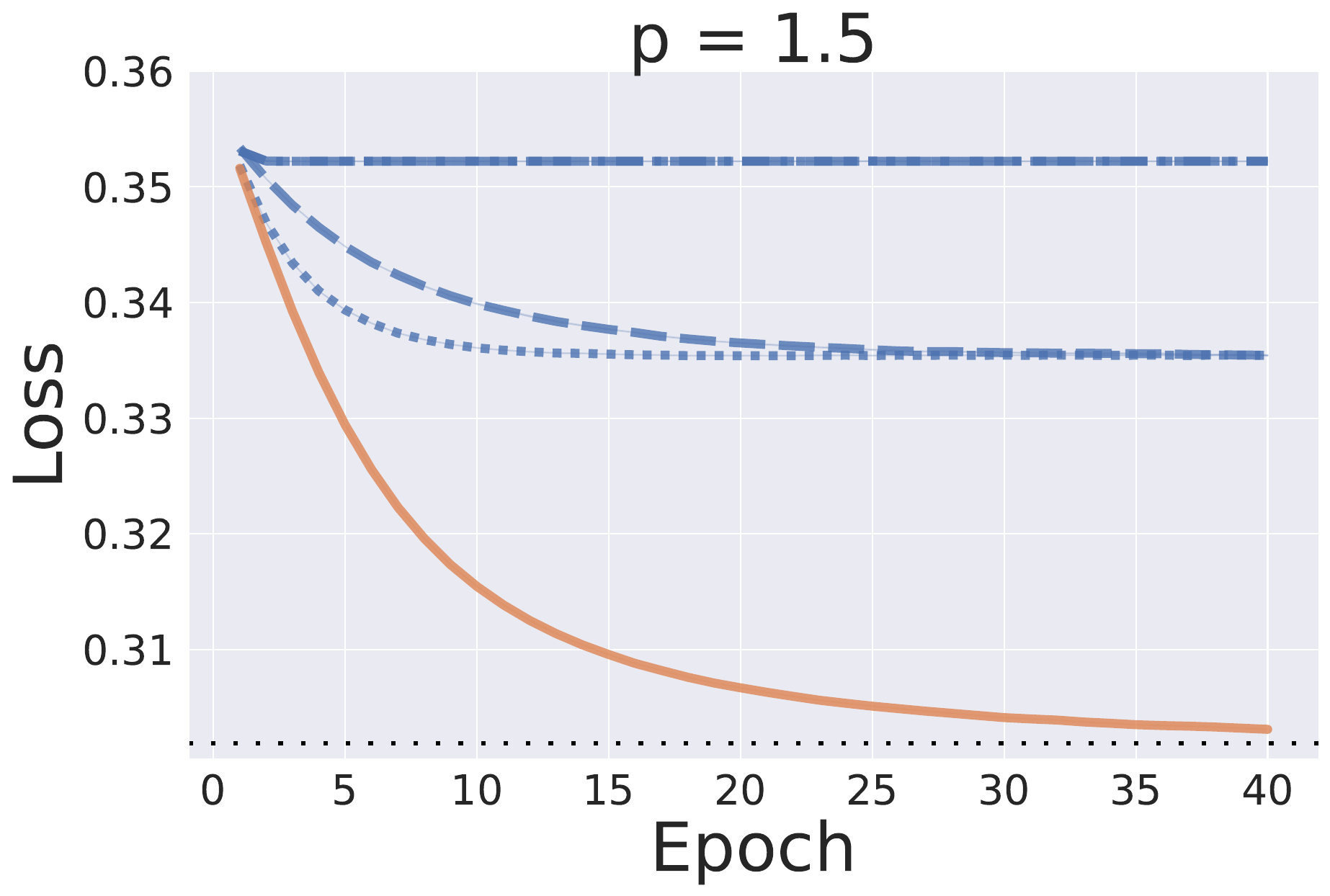}
        \vspace{-4mm}
        \label{}
    \end{subfigure}
    \hfill
    \begin{subfigure}[t]{0.325\linewidth}
        \centering
        \includegraphics[width=1.0\textwidth]{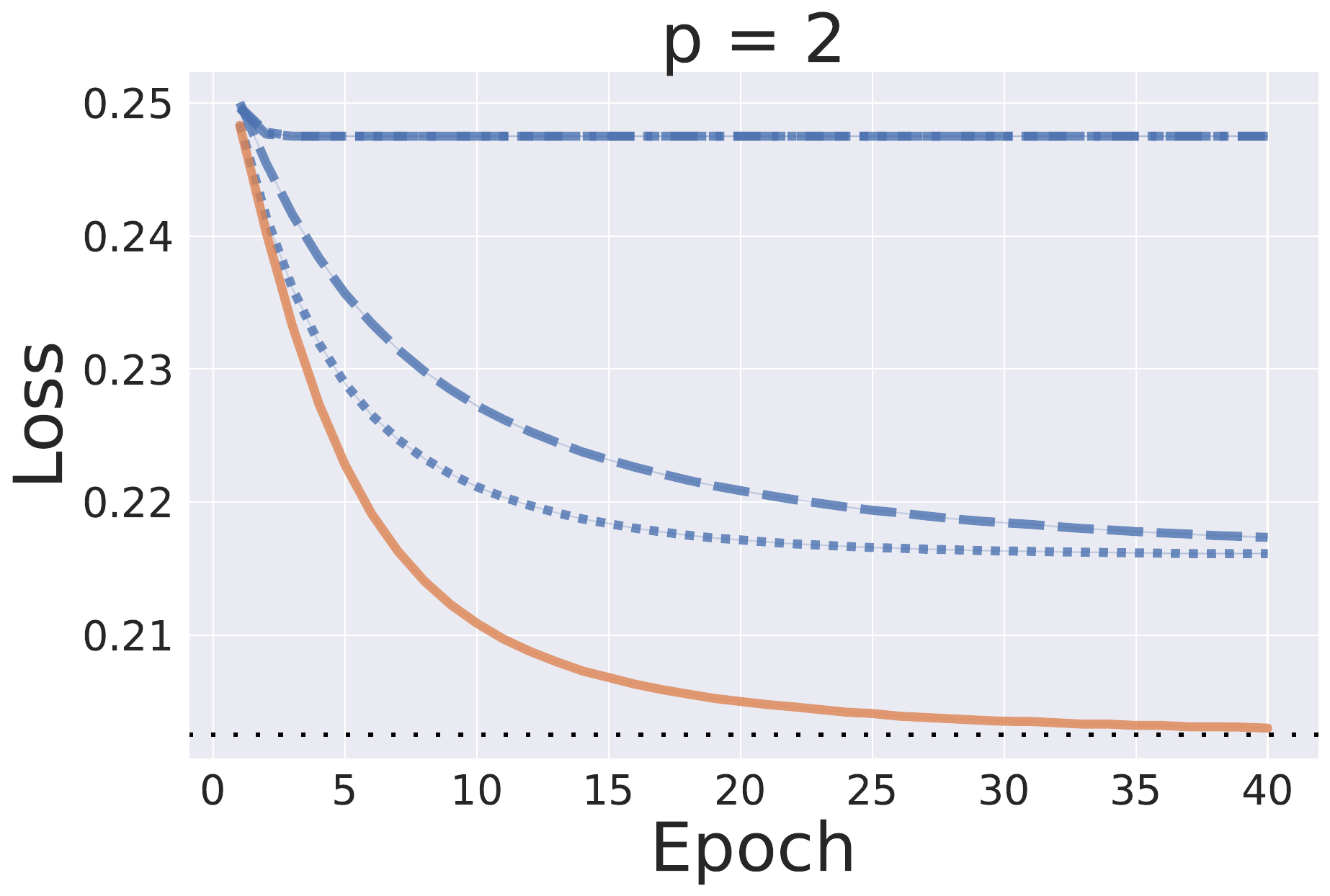}
        \vspace{-4mm}
        \label{}
    \end{subfigure}
    \hfill
    \begin{subfigure}[t]{0.325\linewidth}
        \centering
        \includegraphics[width=1.0\textwidth]{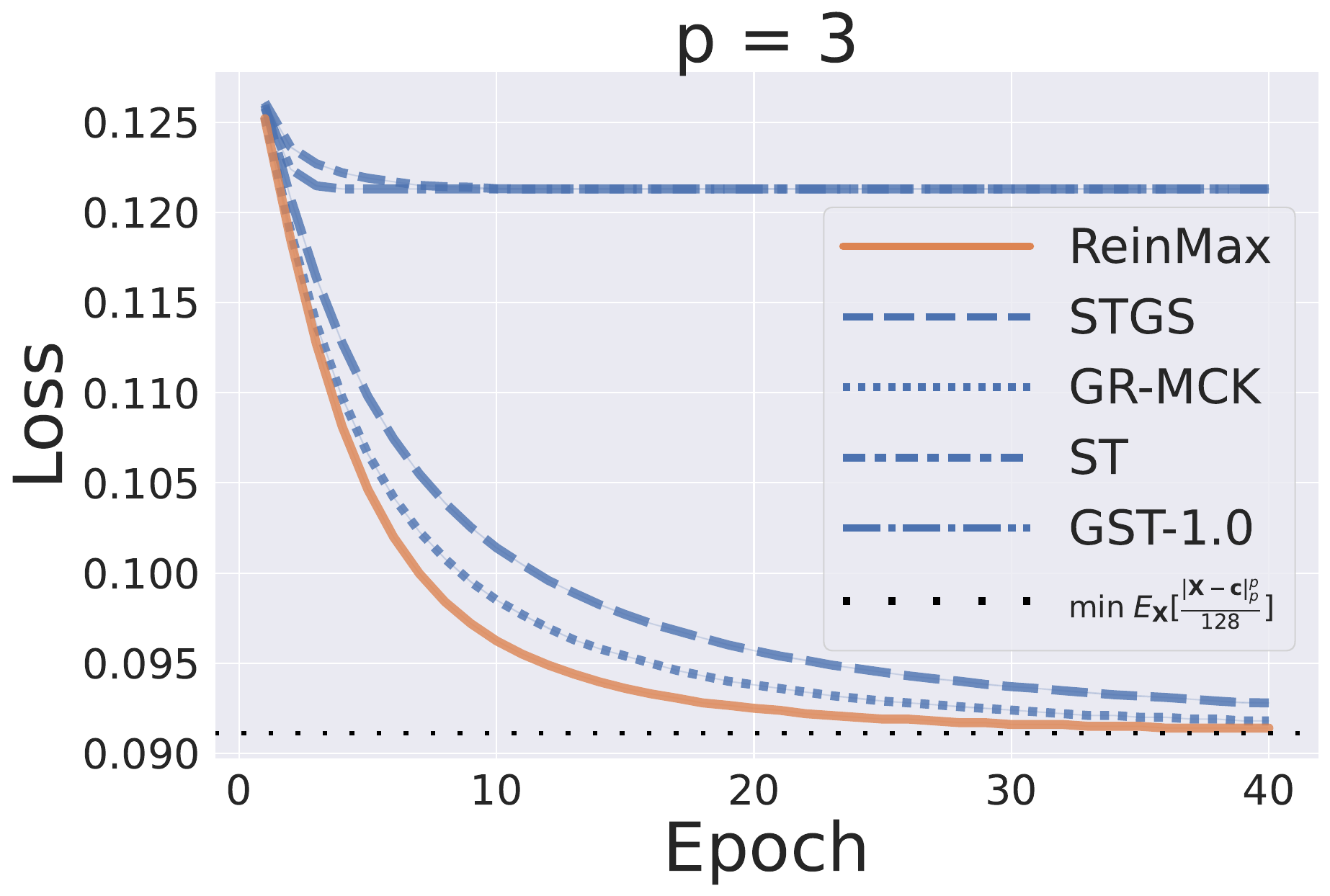} 
        \vspace{-4mm}
        \label{}
    \end{subfigure}
    \vspace{-2.5mm}
    \caption{
    Training curves of polynomial programming, i.e., $\min_\vtheta E_{\mX}\rrrl \|\mX - \vc\|_p^p /128\rrrr, \mbox{ where } \vtheta \in \mathbb{R}^{128\times 2}, \mX \in \{0, 1\}^{128}, \mbox{ and } \mX_i \overset{\mathrm{iid}}{\sim} \mbox{Multinomial}(\mbox{softmax}(\vtheta_i))$.  Details are elaborated in Section~\ref{sect:experiment}.
    }
    \vspace{-4mm}
    \label{fig:poly}
\end{figure}

\begin{minipage}[t]{0.45\textwidth}

\begin{algorithm}[H]
\DontPrintSemicolon
\KwIn{$\vtheta$: softmax input, $\tau$: temperature.}
\KwOut{$\mD$: one-hot samples.}
$\vpi_0 \gets \mbox{softmax}(\vtheta)$ \;
$\mD \gets \mbox{sample\_one\_hot}(\vpi_0)$ \;
$\vpi_1 \gets \mbox{softmax}_\tau(\vtheta)$\;
\tcc{$\mbox{stop\_gradient}(\cdot)$ duplicates its input and detaches it from backpropagation. }
$\mD \gets \vpi_1 - \mbox{stop\_gradient}(\vpi_1) + \mD$ \;
\Return{$\mD$}
\caption{ST. }
\label{algo:st}
\end{algorithm}
\end{minipage}
\hfill
\begin{minipage}[t]{0.52\textwidth}

\begin{algorithm}[H]
\DontPrintSemicolon
\KwIn{$\vtheta$: softmax input,  $\tau$: temperature.}
\KwOut{$\mD$: one-hot samples.}
$\vpi_0 \gets \mbox{softmax}(\vtheta)$ \;
$\mD \gets \mbox{sample\_one\_hot}(\vpi_0)$ \;
$\vpi_1 \gets \frac{\mD + \mbox{softmax}_{\tau}(\vtheta)}{2}$\;
$\vpi_1 \gets \mbox{softmax}\LARGE(\normalsize \mbox{stop\_gradient}(\ln(\vpi_1) - \vtheta) + \vtheta\LARGE)\normalsize$\;
$\vpi_2 \gets 2 \cdot \vpi_1 - \frac{1}{2} \cdot \vpi_0$\;
$\mD \gets \vpi_2 - \mbox{stop\_gradient}(\vpi_2) + \mD$\;
\Return{$\mD$}
\caption{ReinMax. }
\label{algo:reinmax}
\end{algorithm}
\end{minipage} 

First, we adopt a novel perspective to examine ST and show that it works as a special case of the forward Euler method, approximating the gradient with first-order accuracy.
Besides confirming that ST is indeed an approximation of the gradient,  our finding provides guidance on how to optimize hyper-parameters of ST and its variants, i.e., ST prefers to set the temperature $\tau \geq 1$, and Straight-Through Gumbel-Softmax (STGS; \citeauthor{Jang2016CategoricalRW}, \citeyear{Jang2016CategoricalRW}) prefers to set the temperature $\tau \leq 1$. 

Our analyses not only shed insights on the underlying mechanism of ST but also 
lead us to develop a novel gradient estimation method called ReinMax. 
\ours integrates Heun's Method and achieves second-order accuracy, i.e., its approximation matches the Taylor expansion of the gradient to the second order, without requiring the Hessian matrix or other second-order derivatives.  

We conduct extensive experiments 
on polynomial programming~\cite{Tucker2017REBARLU,Grathwohl2017BackpropagationTT,Pervez2020LowBL,Paulus2020RaoBlackwellizingTS}, unsupervised generative modeling~\citep{Kingma2013AutoEncodingVB}, structured output prediction~\citep{Nangia2018ListOpsAD}, and differentiable neural architecture search~\citep{Dong2020NATSBenchBN} to demonstrate that \ours brings consistent improvements over the state of the art\footnote{Implementations are available at 
\url{https://github.com/microsoft/ReinMax}.}. 

Our contributions are two-fold:
\begin{itemize}[leftmargin=*]
    \item
    \vspace{-0.1in}
    We formally establish that ST works as a first-order approximation to the gradient in the general multinomial case, which provides valuable guidance for future research and applications. 
    
    \item 
    We propose a novel and sound gradient estimation method ReinMax that achieves second-order accuracy without requiring the Hessian matrix or other second-order derivatives. ReinMax is shown to outperform the previous state-of-the-art methods in extensive experiments.
\end{itemize}

\section{Related Work and Preliminary}
\label{sect:notation}
\smallsection{Discrete Latent Variables and Gradient Computation}
The idea of incorporating discrete latent variables and neural networks dates back to sigmoid belief network and Helmholtz machines~\citep{Williams1992SimpleSG,Dayan1995TheHM}. 
To keep things straightforward, we will focus on a simplified scenario.
We refer to the tempered softmax as $\mbox{softmax}_\tau (\vtheta)_i = \frac{\exp(\vtheta_i / \tau)}{\sum_{j=1}^n \exp(\vtheta_j / \tau)}$, where $n$ is the number of possible outcomes, $\vtheta \in \mathcal{R}^{n \times 1}$ is the parameter, and $\tau$ is the temperature\footnote{Without specification, the temperature (i.e., $\tau$) is set to $1$.}. 
For $i \in [1, \cdots, n]$, we mark its one-hot representation as $\mI_i \in \mathcal{R}^{n \times 1}$, whose element equals $1$ if it is the $i$-th element or equals $0$ otherwise. 
Let $\mD$ be a discrete random variable and $\mD \in \{\mI_1, \cdots, \mI_n\}$, 
we assume the distribution of $\mD$ is parameterized as: $p(\mD = \mI_i) = \vpi_i = \mbox{softmax}(\vtheta)_i$, and mark $\mbox{softmax}_\tau(\vtheta)$ as $\vpi^{(\tau)}$.
Given a differentiable function $f: \mathcal{R}^{n} \to \mathcal{R}$, we aim to minimize (note that temperature scaling is not used in the generation of $\mD$):
\begin{equation}
    \min_\vtheta \mathcal{L}(\vtheta), \; \mbox{ where } \mathcal{L}(\vtheta)  = E_{\mD \sim \mbox{\small softmax}(\vtheta)} [f(\mD)].
    \label{eqn:objective_as_expectation}
\end{equation}

Here, we mark the gradient of $\vtheta$ as $\nabla$:
\begin{equation}
    \nabla :=  \frac{\partial \mathcal{L}(\vtheta)}{\partial \vtheta} = \sum_i f(\mI_i) \frac{d\,\vpi_i}{d\,\vtheta}.
    \label{eqn:exact}
\end{equation}
In many applications, it is usually too costly to compute $\nabla$, since it requires the computation of $\{f(\mI_1), \cdots, f(\mI_n)\}$ and evaluating $f(\mI_i)$ is costly for typical deep learning applications. 
Correspondingly, many efforts have been made to estimate $\nabla$ efficiently. 

The $\nabla_{\mbox{\scriptsize REINFORCE}}$~\citep{Williams1992SimpleSG} is unbiased (i.e., $E[\nabla_{\mbox{\scriptsize REINFORCE}}]=\nabla$) and only requires the distribution of the discrete variable to be differentiable (i.e., no backpropagation through $f$): 
\begin{equation}
    \nabla_{\mbox{\scriptsize REINFORCE}} := f(\mD) \frac{d\,\log p(\mD)}{d\,\vtheta}.
    \label{eqn:reinforce}
\end{equation}
Despite the REINFORCE estimator being unbiased, it tends to have prohibitively high variance, especially for networks that have other sources of randomness (i.e., dropout or other independent random variables). 
Recently, attempts have been made to reduce the variance of REINFORCE~\citep{Gu2015MuPropUB,Tucker2017REBARLU,Grathwohl2017BackpropagationTT,shi2022gradient}.
Still, it has been found that the REINFORCE-style estimators fail to work well in many real-world applications. 
Empirical comparisons between \ours and REINFORCE-style methods are elaborated in Section~\ref{subsec:exp_reinforce}. 

\smallsection{Efficient Gradient Approximation}
In practice, a popular family of estimators is {Straight-Through (ST) estimators}.
They compute the backpropagation "through" a surrogate that treats the non-differentiable function (e.g., the sampling of $\mD$) as an identity function. 
The idea of ST originates from the perceptron algorithm~\citep{rosenblatt1957perceptron,Mullin1962PrinciplesON}, which leverages a modified chain rule and utilizes the identity function as the proxy of the original derivative of a binary output function. 
\citet{Bengio2013EstimatingOP} improves this method by using non-linear functions like sigmoid or softmax, and \citet{Jang2016CategoricalRW} further incorporates the Gumbel reparameterization.
Here, we briefly describe Straight-Through (ST) and Straight-Through Gumbel-Softmax (STGS). 

In the general multinomial distribution case, as in Algorithm~\ref{algo:st}, the ST estimator treats the sampling process of $\mD$ as an identity function during the backpropagation\footnote{We use the notation $\widehat{\nabla}$ to indicate gradient approximations. Note that the generation of $\mD$ is not differentiable, and $\appnabla_{\mbox{\scriptsize ST}}$ does not have the term $\partial \mD / \partial \vpi$. }: 
\begin{equation}
    \appnabla_{\mbox{\scriptsize ST}} := \frac{\partial f(\mD)}{\partial \mD} \cdot \frac{d\,\vpi }{d\,\vtheta}.
    \label{eqn:st}
\end{equation}
In practice, $\appnabla_{\mbox{\scriptsize ST}}$ is usually implemented with the tempered softmax, under the hope that the temperature hyper-parameter $\tau$ may be able to reduce the bias introduced by $\appnabla_{\mbox{\scriptsize ST}}$~\citep{Chung2016HierarchicalMR}. 

The STGS estimator is built upon the Gumbel re-parameterization trick~\citep{Maddison2014AS, Jang2016CategoricalRW}.  
It is observed that the sampling of $\mD$ can be reparameterized using Gumbel random variables at the zero-temperature limit of the tempered softmax~\citep{Gumbel1954StatisticalTO}: 
\begin{equation}
    \mD = \lim_{\tau \to 0} \mbox{softmax}_\tau (\vtheta + \mG) \;\; \mbox{where } \mG_i \mbox{ are i.i.d. and } \mG_i \sim \mbox{Gumbel}(0, 1).
\nonumber
\end{equation}
STGS treats the zero-temperature limit as identity function during the backpropagation:
\begin{equation}
    \appnabla_{\mbox{\scriptsize STGS}} := \frac{\partial f(\mD)}{\partial \mD} \cdot \frac{d\,\mbox{softmax}_\tau (\vtheta + \mG) }{ d\,\vtheta}.
    \label{eqn:stgs}
\end{equation}

Both $\appnabla_{\mbox{\scriptsize ST}}$ and $\appnabla_{\mbox{\scriptsize STGS}}$ are clearly biased. 
However, since the mechanism of ST is unclear, it remains unanswered what the form of their biases are, how to configure their hyper-parameters for optimal performance, or even whether $E[\appnabla_{\mbox{\scriptsize ST}}]$ or $E[\appnabla_{\mbox{\scriptsize STGS}}]$ can be viewed as an approximation of $\nabla$. 
Thus, we aim to answer the following questions: \emph{How $\appnabla_{\mbox{\scriptsize ST}}$ approximates $\nabla$ and how it can be improved?}

\section{Discrete Variable Gradient Approximation: a Numerical ODE Perspective}
\label{sect:first-order}

In numerical analysis, extensive studies have been conducted to develop numerical methods for solving ordinary differential equations. 
In this study, we leverage these methods to approximate $\nabla$ with the gradient of $f$.
To begin, we demonstrate that ST works as a first-order approximation of $\nabla$.
Then, we propose ReinMax, which integrates Heun's method for a better gradient approximation and achieves second-order accuracy. 

\subsection{Straight-Through as a First-order Approximation}
\label{subsec: st}

We start by defining a first-order approximation of $\nabla$ as $\appnabla_{\mbox{\small 1st-order}}$. 
\begin{definition}
\label{def:1st}
One first-order approximation of $\nabla$ is
$\appnabla_{\mbox{\small 1st-order}} :=  \sum_i \sum_j  \vpi_j \frac{\partial f(\mI_j)}{\partial \mI_j} (\mI_i - \mI_j) \frac{d\,\vpi_i}{d\,\vtheta}$.
\end{definition}
To understand why $\appnabla_{\mbox{\small 1st-order}}$ is a first-order approximation, we rewrite $\nabla$ in Equation~\ref{eqn:exact} as\footnote{Please note that $\sum_i E[f(\mD)] \frac{d\,\vpi_i}{d\,\vtheta} = E[f(\mD)] \frac{d\,\sum_i \vpi_i}{d\,\vtheta}=E[f(\mD)] \frac{d \bm{1}}{d\,\vtheta}=0$}:
\begin{eqnarray}
    \nabla =  \sum_i (f(\mI_i) - E[f(\mD)]) \frac{d\,\vpi_i}{d\,\vtheta} + \sum_i E[f(\mD)] \frac{d\,\vpi_i}{d\,\vtheta}= \sum_i \sum_j  \vpi_j (f(\mI_i) - f(\mI_j)) \frac{d\,\vpi_i}{d\,\vtheta}.
    \label{eqn:exact_baseline_final}
\end{eqnarray}
Comparing $\appnabla_{\mbox{\small 1st-order}}$ and Equation~\ref{eqn:exact_baseline_final}, it is easy to notice that $\appnabla_{\mbox{\small 1st-order}}$ approximates $f(\mI_i) - f(\mI_j)$ as $ \frac{\partial f(\mI_j)}{\partial \mI_j} (\mI_i - \mI_j)$. 
In numerical analyses, this approximation is known as the forward Euler method, which has first-order accuracy (we provide a brief introduction to the forward Euler method in Appendix~\ref{appendix:ode}). 
Correspondingly, we know that $\appnabla_{\mbox{\small 1st-order}}$ is a first-order approximation of $\nabla$. 

Now, we proceed to show $\appnabla_{\mbox{\scriptsize ST}}$ works as a first-order approximation. Note that our analyses only apply to $\appnabla_{\mbox{\scriptsize ST}}$ as defined in Equation~\ref{eqn:st} and may not apply to its other variants.  
\begin{theorem}
\begin{eqnarray}
    E[\appnabla_{\mbox{\scriptsize ST}}] = \appnabla_{\mbox{\small 1st-order}}. 
    \nonumber
\end{eqnarray}
\label{theorem: st}
\end{theorem}
The proof of Theorem~\ref{theorem: st} is provided in Appendix~\ref{appendix:proof-st}. 

It is worth mentioning that \cite{Tokui2017EvaluatingTV} discussed this connection for the special case of $\mD$ being a Bernoulli variable. 
However, their study is built upon a Bernoulli variable property (i.e., $\nabla = (f(\mI_2) - f(\mI_1)) \frac{d \vpi_1 }{d \theta} = (f(\mI_1) - f(\mI_2)) \frac{d \vpi_2 }{d \theta}$), making their analyses not applicable to multinomial variables. 
Alternatively, the analyses in \citet{Gregor2013DeepAN} and \citet{Pervez2020LowBL} are applicable to multinomial variables but resort to  modify $\appnabla_{\mbox{\scriptsize ST}}$ as $\frac{1}{n\cdot \vpi_\mD}\appnabla_{\mbox{\scriptsize ST}}$, in order to position it as a first-order approximation. 
We suggest that this modification would lead to unwanted instability and provide more discussions in Section~\ref{subsec: altervative-first-order} and Section~\ref{subsec: exp-discussions}.  
Here, our study is the first to formally established  $\appnabla_{\mbox{\scriptsize ST}}$ works as a first-order approximation in the general multinomial case.

Besides revealing the mechanism of the Straight-Through estimator, our finding also shows that the bias of $\appnabla_{\mbox{\scriptsize ST}}$ comes from using the first-order approximation (i.e., the forward Euler method). 
Accordingly, we propose to integrate a better approximation for $f(\mI_i) - f(\mI_j)$.

\subsection{Towards Second-order Accuracy: \ours}
\label{subsect:second-order}

The literature on numerical methods for differential equations shows that it is possible to achieve higher-order accuracy \emph{without computing higher-order derivatives}.
Correspondingly, we propose to integrate a second-order approximation to reduce the bias of the gradient estimator. 
\begin{definition}
One second-order approximation of $\nabla$ is
$$
\appnabla_{\mbox{\small 2nd-order}} :=  \sum_i \sum_j  \frac{\vpi_j }{2} (\frac{\partial f(\mI_j)}{\partial \mI_j} + \frac{\partial f(\mI_i)}{\partial \mI_i} ) (\mI_i - \mI_j) \frac{d\,\vpi_i}{d\,\vtheta}.
$$
\end{definition}
Comparing $\appnabla_{\mbox{\small 2nd-order}}$ and Equation~\ref{eqn:exact_baseline_final}, we can observe that,  $\appnabla_{\mbox{\small 2nd-order}}$ approximates $f(\mI_i) - f(\mI_j)$ as $\frac{1}{2} (\frac{\partial f(\mI_i)}{\partial \mI_i} + \frac{\partial f(\mI_j)}{\partial \mI_j})(\mI_i - \mI_j)$. 
This approximation is known as the Heun's Method and has second-order accuracy (we provide a brief introduction to Heun's method in Appendix~\ref{appendix:ode}). 
Correspondingly, we know that $\appnabla_{\mbox{\small 2nd-order}}$ is a second-order approximation of $\nabla$. 

Based on this approximation, we propose the ReinMax operator as ($\vpi_\mD$ refers to $\frac{\vpi + \mD}{2}$, $\tI$ refers to the identity matrix, and $\odot$ refers to the element-wise product):
\begin{equation}
\appnabla_{\mbox{\scriptsize ReinMax}} := 2 \cdot \appnabla^{\frac{\vpi + \mD}{2}}- \frac{1}{2}\appnabla_{\mbox{\scriptsize ST}}, \mbox{ where } \appnabla^{\frac{\vpi + \mD}{2}} = \frac{\partial f(\mD)}{\partial \mD} \cdot \large(\normalsize (\vpi_\mD \cdot 1^T) \odot \tI - \vpi_\mD \cdot \vpi_\mD^T\large)\normalsize
\label{eqn:reinmax}
\end{equation}
Then, we show that $\appnabla_{\mbox{\scriptsize ReinMax}}$ approximates $\nabla$ to the second order. Or, formally we have:
\begin{theorem}
\begin{eqnarray}
    E[\appnabla_{\mbox{\scriptsize ReinMax}}] = \appnabla_{\mbox{\small 2nd-order}}.
    \nonumber
\end{eqnarray}
\label{theorem: reinmax}
\end{theorem}
The proof of Theorem~\ref{theorem: reinmax} is provided in Appendix~\ref{appendix:proof-reinmax}.

\smallsection{Computation Efficiency of ReinMax}
Instead of requiring Hessian or other second-order derivatives, 
$\appnabla_{\mbox{\scriptsize ReinMax}}$ achieves second-order accuracy with two first-order derivatives (i.e., $\frac{\partial f(\mI_j)}{\partial \mI_j}$ and $\frac{\partial f(\mI_i)}{\partial \mI_i}$).
As observed in our empirical efficiency comparisons in Section~\ref{sect:experiment}, the computation overhead of $\appnabla_{\mbox{\scriptsize ReinMax}}$ is negligible.  
At the same time, similar to $\appnabla_{\mbox{\scriptsize ST}}$ (as in Algorithm~\ref{algo:st}), our proposed algorithm can be easily integrated with existing automatic differentiation toolkits like PyTorch (a simple implementation of \ours is provided in Algorithm~\ref{algo:reinmax}), making it easy to be integrated with existing algorithms. 

\smallsection{Applicability of Higher-order ODE solvers}
Although it's possible to apply higher-order ODE solvers, they require more gradient evaluations, leading to undesirable computational overhead. To illustrate this point:
The approximation used by ReinMax requires n gradient evaluations, i.e., $\{\frac{\partial f(\mI_i)}{\partial \mI_i}\}$.
In contrast, the approximation derived by RK4 needs $n^2+n$ gradient evaluations, i.e., $\{\frac{\partial f(\mI_i)}{\partial \mI_i}\}$ and $\{\frac{\partial f(\mI_{ij})}{\partial \mI_{ij}}\}$, where $\mI_{ij} = \frac{\mI_i + \mI_j}{2}$.
Therefore, while higher-order solvers are applicable, they may not be suitable in our case.

\section{\ours and Baseline Subtraction}
\label{subsec: baseline}

Equation~\ref{eqn:exact_baseline_final} plays a crucial role in positioning ST as a first-order approximation of the gradient and deriving our proposed method, ReinMax.
This equation is commonly referred to as baseline subtraction, a common technique for reducing the variance of REINFORCE. 

In this section, we first discuss the reason for choosing $E[f(\mD)]$ as the baseline, and then reveal that the derivation of \ours is independent to baseline subtraction.

\subsection{Benefits of Choosing $E[f({\mathbf{D}})]$ as the Baseline}
\label{subsec: altervative-first-order}

The choice of baseline in reinforcement learning has been the subject of numerous discussions~\citep{Weaver2001TheOR,Rennie2016SelfCriticalST,shi2022gradient}. 
Similarly, in our study, different baselines lead to different gradient approximations. 

Here, we discuss the rationale for choosing $E[f(\mD)]$ as the baseline.
Considering $\sum_i \vphi_i f(\mI_i)$ as the general form of the baseline ($\vphi_i$ is a distribution over $\{\mI_1, \cdots, \mI_n\}$, i.e., $\sum_i \vphi_i = 1$), we have:
\vspace{1mm}
\begin{remark}
When $\sum_i \vphi_i f(\mI_i)$ is used as the baseline and  $f(\mI_i) - f(\mI_j)$ is approximated as $\frac{\partial f(\mI_j)}{\partial \mI_j} (\mI_i - \mI_j)$, we mark the resulting first-order approximation of $\nabla$ as $\appnabla_{\mbox{\small 1st-order-avg-baseline}}$.
Then, we have
$E[\frac{\vphi_\mD}{\vpi_\mD}\appnabla_{\mbox{\scriptsize ST}}] = \appnabla_{\mbox{\small 1st-order-avg-baseline}}.$
\label{theorem: st-avg}
\end{remark}
The derivations of Remark~\ref{theorem: st-avg} are provided in Appendix~\ref{appendix:proof-st-avg}. 
Intuitively, since $\vpi_\mD$ is the output of the softmax function, it could have very small values, which makes $\frac{\vphi_\mD}{\vpi_\mD}$ to be unreasonably large and leads to undesired instability. 
Therefore, we suggest that $E[f(\mD)]$ is a better choice of baseline when it comes to gradient approximation, since its corresponding gradient approximation is free of the instability $\frac{\vphi_\mD}{\vpi_\mD}$ brought.

It is worth mentioning that, when setting $\vphi$ as $\frac{1}{n}$, the result of Remark~\ref{theorem: st-avg} echoes some existing studies. 
Specifically, both \citet{Gregor2013DeepAN} and \citet{Pervez2020LowBL} propose to approximate $\nabla$ as E[$\frac{1}{n \cdot \vpi_\mD}\appnabla_{\mbox{\scriptsize ST}}$], which matches the result of Remark~\ref{theorem: st-avg} by setting $\vphi = \frac{1}{n}$. 

In Section~\ref{sect:experiment}, we compared the corresponding second-order approximation when treating $E[f(\mD)]$ and $\frac{1}{n} \sum_i f(\mI_i)$ as the baseline, respectively. 
We observed that gradient estimators that use $E[f(\mD)]$ as the baseline consistently outperform gradient estimators that use $\frac{1}{n} \sum_i f(\mI_i)$ as the baseline, which verifies our intuition and demonstrates the importance of the baseline selection.

\subsection{Independence of ReinMax over Baseline Subtraction}

To better understand the effectiveness of ReinMax, we further provide an alternative derivation that does not rely on the selection of the baseline. 
For simplicity, we only discuss $\frac{\partial \mathcal{L}}{\partial \vtheta_k}$ and mark it as $\nabla_k$. 
Similar to Equation~\ref{eqn:exact}, we have: 
\begin{eqnarray}
    \nabla_{k} := \frac{\partial \mathcal{L}}{\partial \vtheta_k} = \sum_i f(\mI_i) \frac{d\,\vpi_i}{d\,\vtheta_k} = \vpi_k \sum_i \vpi_i (f(\mI_k) - f(\mI_i)).
    \label{eqn: exact-nobaseline}
\end{eqnarray}
It is worth mentioning that the derivation of Equation~\ref{eqn: exact-nobaseline} leverages the 
derivative
of the softmax function (i.e., for $\vpi = \mbox{softmax}(\vtheta)$, we have $\partial \vpi_i / \partial \vtheta_k = \vpi_k (\delta_{ik} - \vpi_i)$) and does not involve the baseline subtraction technology. 

\begin{remark}
In Equation~\ref{eqn: exact-nobaseline}, we approximate $f(\mI_k) - f(\mI_i)$ as $\frac{1}{2} (\frac{\partial f(\mI_i)}{\partial \mI_i} + \frac{\partial f(\mI_k)}{\partial \mI_k} )(\mI_k - \mI_i)$, and mark the resulting second-order approximation of $\nabla_k$ as $\appnabla_{\mbox{\small 2nd-order-wo-baseline}, k} =  \vpi_k \sum_i \vpi_i \frac{1}{2} (\frac{\partial f(\mI_i)}{\partial \mI_i} + \frac{\partial f(\mI_k)}{\partial \mI_k} ) (\mI_k - \mI_i)$,
Then, we have $E[\appnabla_{\mbox{\scriptsize ReinMax}}] = \appnabla_{\mbox{\small 2nd-order-wo-baseline}}$
\label{theorem: reinmax-nobaseline}
\end{remark}
The proof of Remark~\ref{theorem: reinmax-nobaseline} is provided in Appendix~\ref{appendix:proof-reinmax-nobaseline}.

As in Remark~\ref{theorem: reinmax-nobaseline}, applying the Heun's method on Equation~\ref{eqn: exact-nobaseline} and Equation~\ref{eqn:exact_baseline_final} lead to the same gradient estimator, 
which implies another benefit of using  $E[f(\mD)]$ as the baseline: the resulting gradient estimator does not rely on additional prior (i.e., its derivation can be free of baseline subtraction).

\begin{figure}[t!]
    \centering
    \includegraphics[width=1.\textwidth]{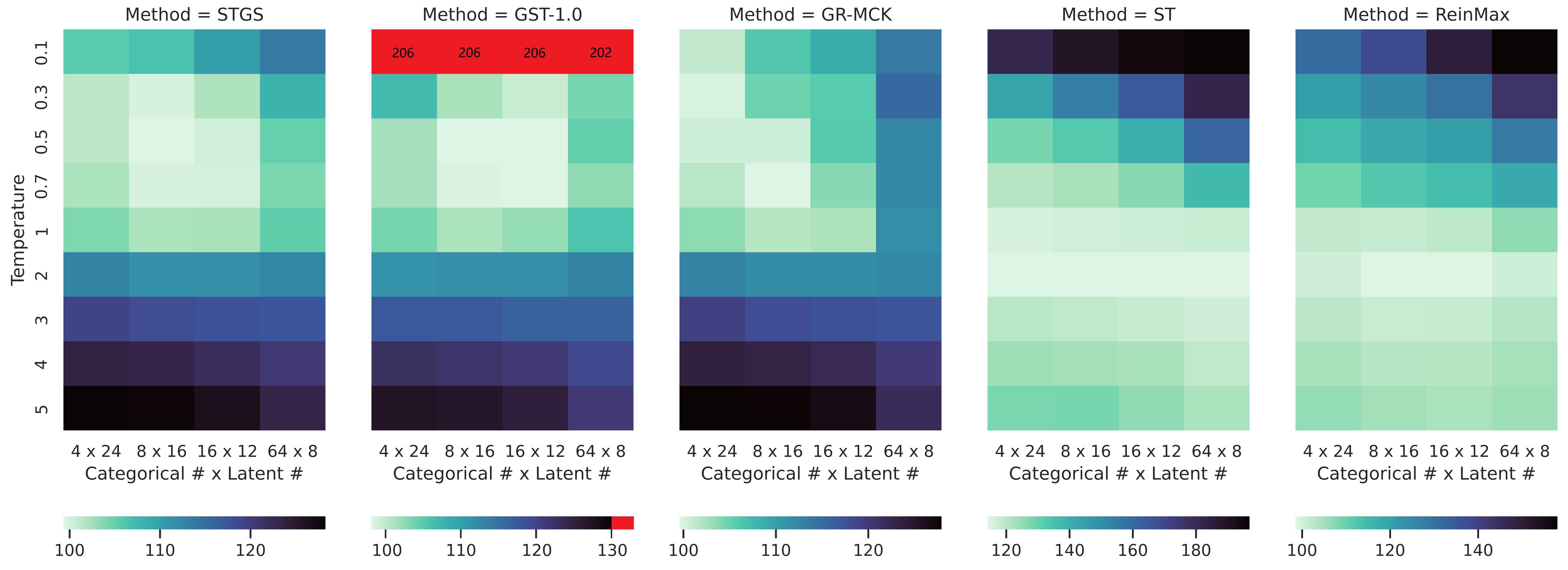}
    \vspace{-6mm}
    \caption{Training $-$ELBO on MNIST-VAE (lighter color indicates better performance). STGS, GST-1.0, and GR-MCK prefer to set the temperature $\tau \leq 1$. ST and ReinMax prefer to set $\tau \geq 1$. }
    \vspace{-3mm}
    \label{fig:temperature}
\end{figure}

\section{Temperature Scaling for Gradient Estimators}
\label{sec:temperature}

Here, we discuss how to apply temperature scaling, a technique widely used in gradient estimators, to our proposed method, ReinMax. 
While the typical practice
is to set the temperature $\tau$ to small values for STGS, we show that ST and ReinMax need a different strategy.

\smallsection{Temperature Scaling for $\nabla_{\mbox{\scriptsize STGS}}$}
As introduced in Section~\ref{sect:notation}, $\nabla_{\mbox{\scriptsize STGS}}$ conduct a two-step approximation: (1) it approximates $\min_\theta E[f(\mD)]$ as $\min_\theta E[f(\mbox{\small softmax}_\tau (\vtheta+\mG)))]$; (2) it approximates $\frac{\partial f(\mbox{\small softmax}_\tau (\vtheta+\mG))}{ \partial \mbox{\small softmax}_\tau (\vtheta+\mG)}$ as $\frac{f(\mD)}{ \partial \mD}$. 
Since the bias introduced in both steps can be controlled by $\tau$, $\nabla_{\mbox{\scriptsize STGS}}$ prefers to set $\tau$ as a relatively small value. 

\smallsection{Temperature Scaling for $\nabla_{\mbox{\scriptsize ST}}$ and $\nabla_{\mbox{\scriptsize ReinMax}}$}
As in Section~\ref{subsec: baseline}, it does not involve temperature scaling to show $\nabla_{\mbox{\scriptsize ST}}$ and $\nabla_{\mbox{\scriptsize ReinMax}}$ work as the first-order and the second-order approximation to the gradient. 
Correspondingly, temperature scaling technology cannot help to reduce the bias for $\nabla_{\mbox{\scriptsize ST}}$ in the same way it does for $\nabla_{\mbox{\scriptsize STGS}}$. 
As in Figure~\ref{fig:temperature}, STGS, GR-MCK, and GST-1.0 work better when setting the temperature $\tau \leq 1$. 
ST and ReinMax work better when setting the temperature $\tau \geq 1$.

Thus, we incorporate temperature scaling to smooth the gradient approximation ($\vpi_\tau = \mbox{softmax}_\tau (\vtheta)$) as
$\appnabla_{\mbox{\scriptsize ReinMax}} = 2 \cdot \appnabla^{\frac{\vpi_\tau + \mD}{2}} - \frac{1}{2} \appnabla_{\mbox{\scriptsize ST}}.$
It is worth emphasizing that $\tau$ in $\appnabla_{\mbox{\scriptsize ReinMax}}$ is used to stabilize the gradient approximation (instead of reducing bias) at the cost of accuracy. 
Therefore, the value of $\tau$ should be larger or equal to $1$. 

\section{Experiments}
\label{sect:experiment}

Here, we conduct experiments on polynomial programming, unsupervised generative modeling, and structured output prediction. 
In all experiments, we consider four major baselines: Straight-Through (ST), Straight-Through Gumbel-Softmax (STGS), Gumbel-Rao Monte Carlo (GR-MCK), and Gapped Straight-Through (GST-1.0). 
For a more comprehensive comparison, we run a complete grid search on the training hyper-parameters for all methods. 
Also, we would reference results from the literature when their setting is comparable with ours. 
More details are elaborated in Appendix~\ref{appendix:exp}.

\begin{figure}[b]
    \centering
    \vspace{-5mm}
    \begin{subfigure}[t]{0.45\linewidth}
        \centering
        \includegraphics[width=1.1\textwidth]{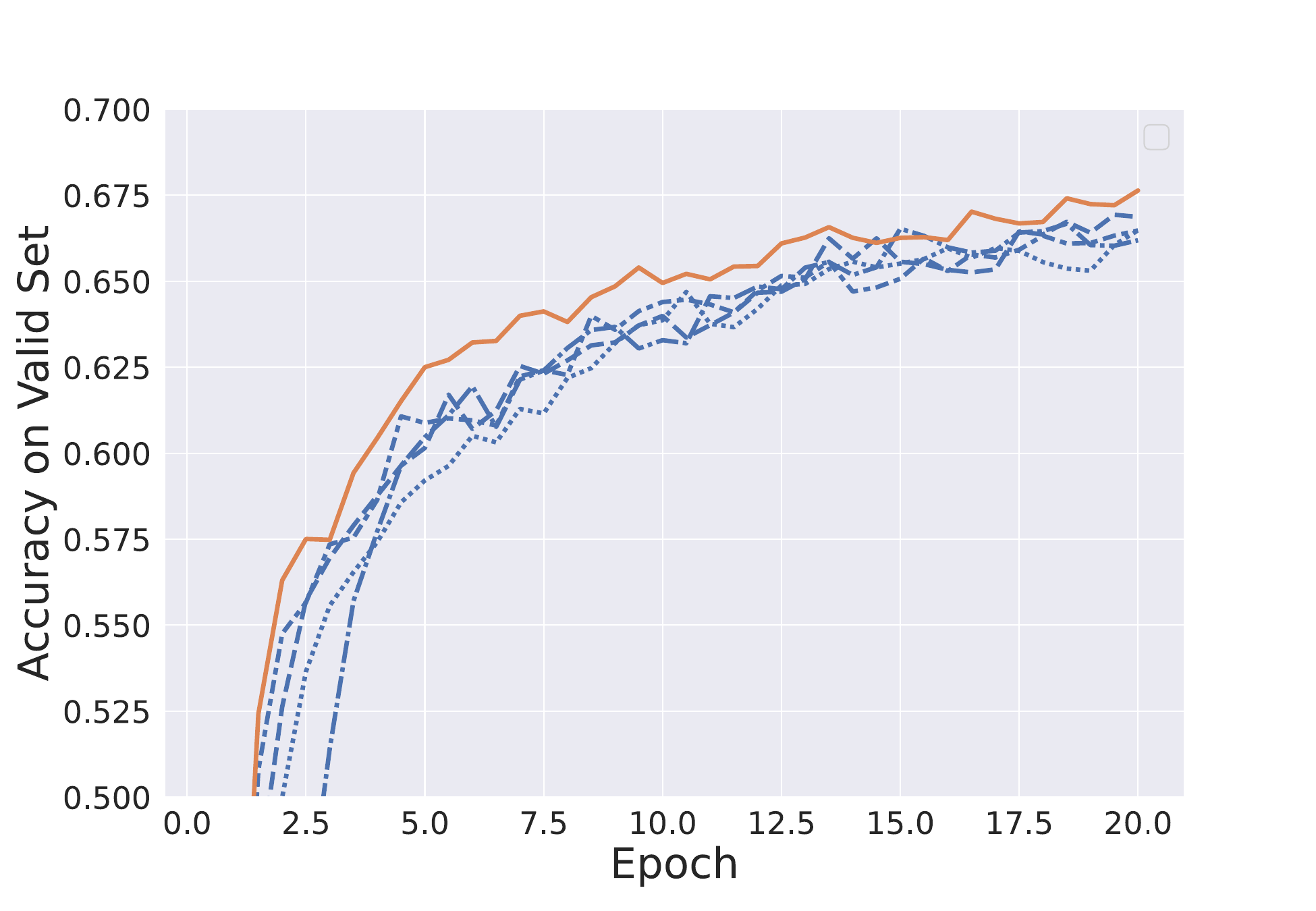}
        \vspace{-4mm}
        \label{}
    \end{subfigure}
    \begin{subfigure}[t]{0.45\linewidth}
        \centering
        \includegraphics[width=1.1\textwidth]{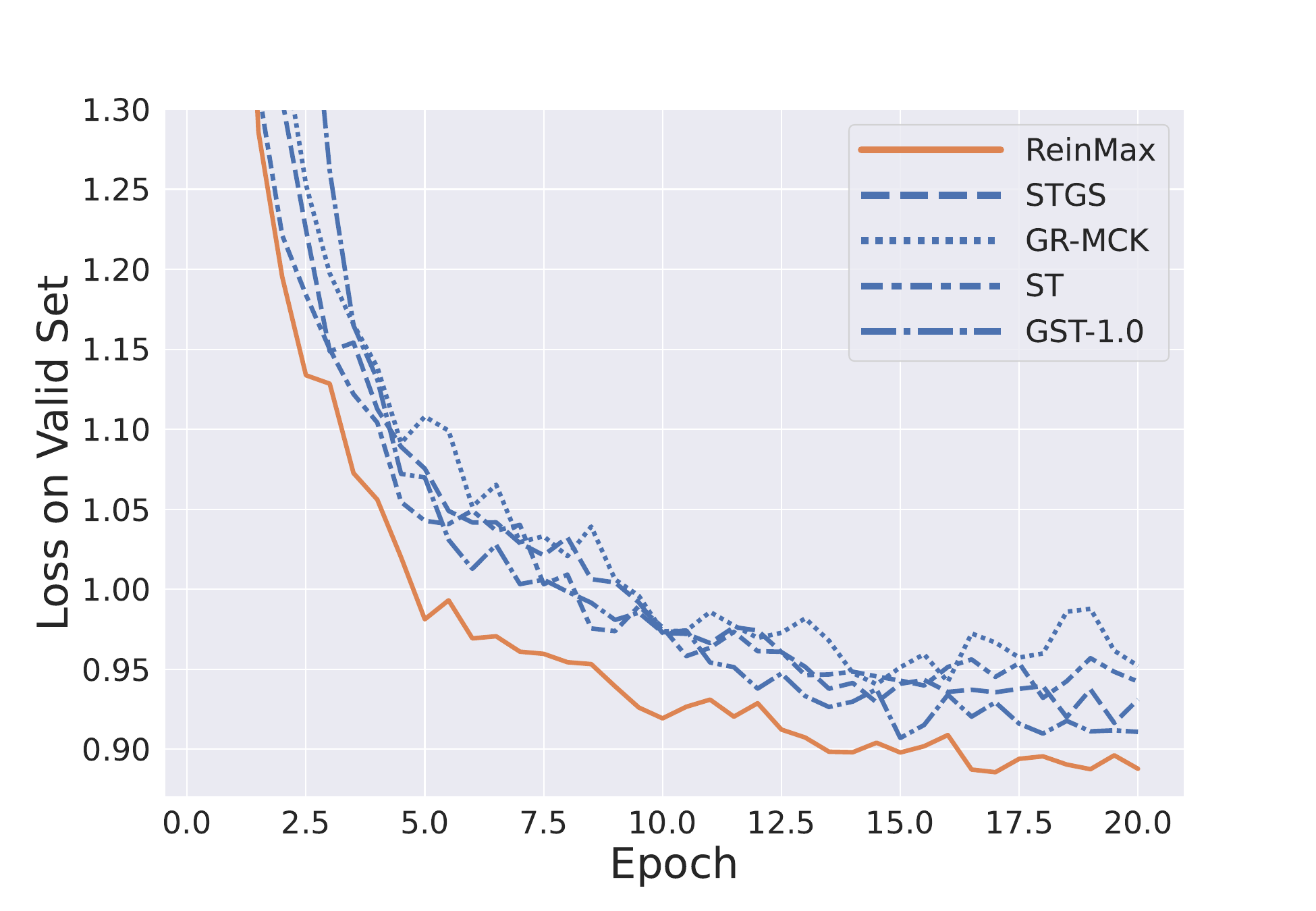}
        \vspace{-4mm}
        \label{}
    \end{subfigure}
    \vspace{-3mm}
    \caption{
    The accuracy (left) and loss (right) on the valid set of ListOps.  
    }
    \label{fig:listops}
    \vspace{-4mm}
\end{figure}

\subsection{Polynomial Programming}

Following previous studies~\citep{Tucker2017REBARLU,Grathwohl2017BackpropagationTT,Pervez2020LowBL,Paulus2020RaoBlackwellizingTS}, we start with a simple problem. 
Consider $L$ i.i.d. latent binary variables $\mX_1, \cdots, \mX_L \in \{0, 1\}$ and a constant vector $\vc \in \mathcal{R}^{L \times 1}$, we parameterize the distributions of $\{\mX_1, \cdots, \mX_L\}$ with $L$ softmax functions, i.e., $\mX_i \overset{\mathrm{iid}}{\sim}  \mbox{Multinomial}(\mbox{softmax}(\vtheta_i))$ and $\vtheta_i \in \mathcal{R}^2$. 
Following previous studies, we set every dimension of $\vc$ as $0.45$, i.e., $\forall i, \vc_i = 0.45$, and use $\min_\vtheta E_{\mX} \rrrl\frac{\|\mX - \vc\|_p^p}{L}\rrrr$ as the objective. 

\smallsection{Training Curve with Various $p$}
We first set the number of latent variables (i.e., $L$) as 128 and batch size as 256. 
The training curve is visualized in Figure~\ref{fig:poly} for $p = 1.5$, $2$, and $3$. 
In all cases, \ours achieved near-optimal performance and the best convergence speed.
Meanwhile, we can observe that ST and GST-1.0 do not perform well in all three cases.
Although the final performance of STGS and GR-MCK is close to ReinMax, ReinMax has a faster convergence speed.

\begin{table}[t!]

\centering
\caption{Performance on ListOps.}
\label{table:listops}
\scalebox{0.9}{
\begin{tabular}{lccccc}
\toprule
  & STGS & GR-MCK & GST-1.0 & ST & \ours \\ 
\midrule
Valid Accuracy & 66.95$\pm$3.05 &  66.53$\pm$0.58 & 66.28$\pm$0.52  & 66.51$\pm$0.76  & \textbf{67.65$\pm$1.25} \\
\midrule
Test Accuracy & 67.30$\pm$2.50 &  66.53$\pm$0.86 & 66.30$\pm$0.62  & 66.26$\pm$0.48  & \textbf{68.07$\pm$1.18} \\
\bottomrule
\end{tabular}
}
\vspace{-0.3cm}
\end{table}

\begin{table}[t!]

\centering
\caption{Training $-$ELBO on MNIST ($N\times M$ refers to $N$ categorical dim. and $M$ latent dim.). }
\label{table:mnist-vae}
\scalebox{.81}{
\begin{tabular}{l|c|c|cccc|c}
\toprule
  & AVG & $8\times 4$ & $4\times 24$ & $8\times 16$ & $16\times 12$ & $64\times 8$ & $10\times 30$ \\ 
\midrule
STGS    & 105.20 & 126.85$\pm$0.85 & 101.32$\pm$0.43 &  99.32$\pm$0.33 & 100.09$\pm$0.32 & 104$\pm$0.41 &  99.63$\pm$0.63 \\ 
GR-MCK  & 107.06 & 125.94$\pm$0.71 &  99.96$\pm$0.25 &  99.58$\pm$0.31 & 102.54$\pm$0.48 & 112.34$\pm$0.48 & 102.02$\pm$0.18 \\ 
GST-1.0 & 104.25 & 126.35$\pm$1.24 & 101.49$\pm$0.44 &  98.29$\pm$0.66 &  98.12$\pm$0.57 & 102.53$\pm$0.57 &  98.64$\pm$0.33 \\ 
\midrule
ST      & 116.72 & 135.53$\pm$0.31 & 112.03$\pm$0.03 & 112.94$\pm$0.32 & 113.31$\pm$0.43 & 113.90$\pm$0.28 & 112.63$\pm$0.34 \\
\ours   & \textbf{103.21} & \textbf{124.66$\pm$0.88} &  \textbf{99.77$\pm$0.45} & \textbf{97.70$\pm$0.39} & \textbf{98.06$\pm$0.53} & \textbf{100.71$\pm$0.70} &  \textbf{98.37$\pm$0.44} \\
\bottomrule
\end{tabular}
}
\vspace{-3mm}
\end{table}

\begin{figure}[t!]
    \centering
    \vspace{-10mm}
    \begin{subfigure}[t]{0.45\linewidth}
        \centering
        \includegraphics[width=1.1\textwidth]{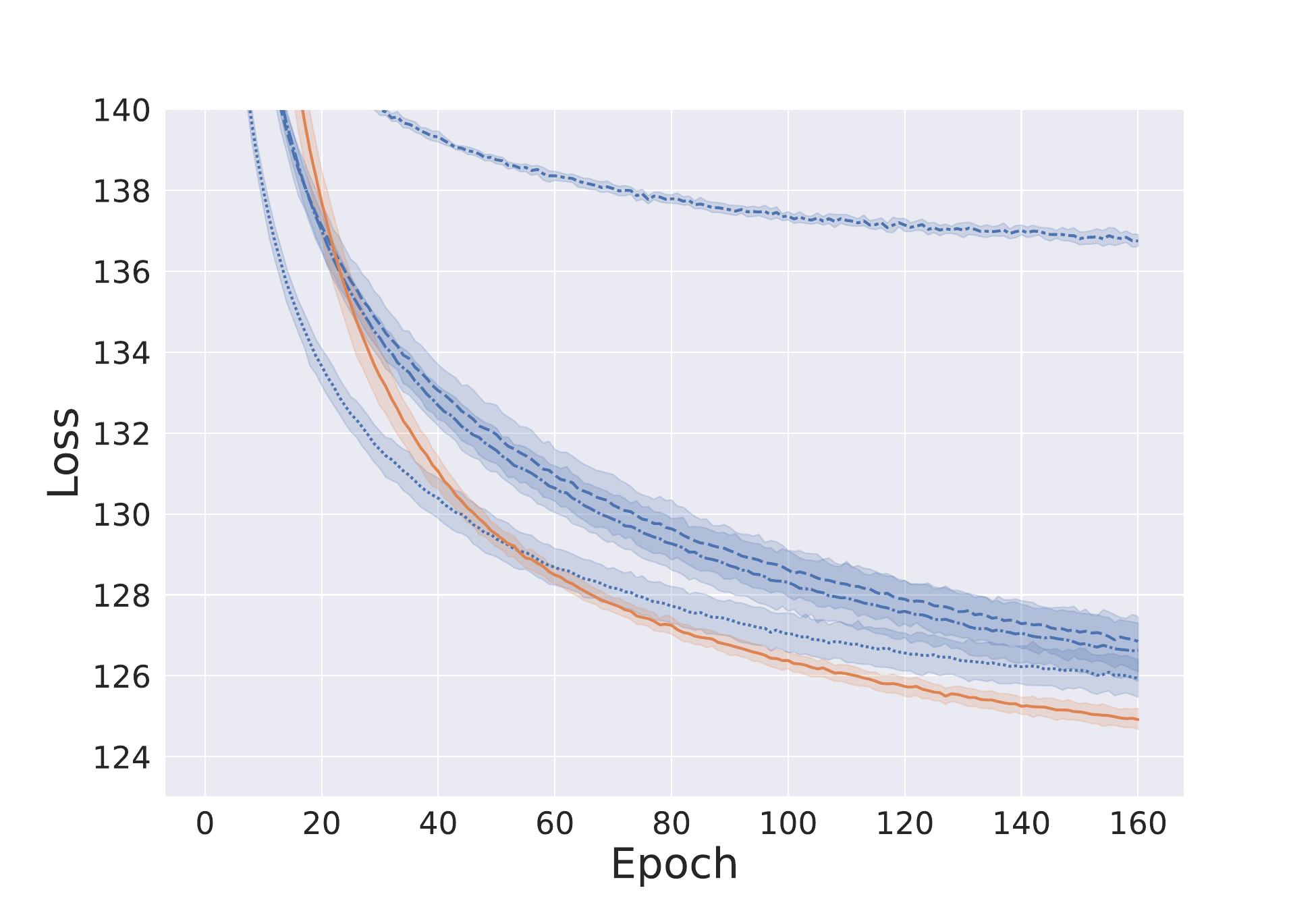}
        \vspace{-4mm}
        \label{}
    \end{subfigure}
    \begin{subfigure}[t]{0.45\linewidth}
        \centering
        \includegraphics[width=1.1\textwidth]{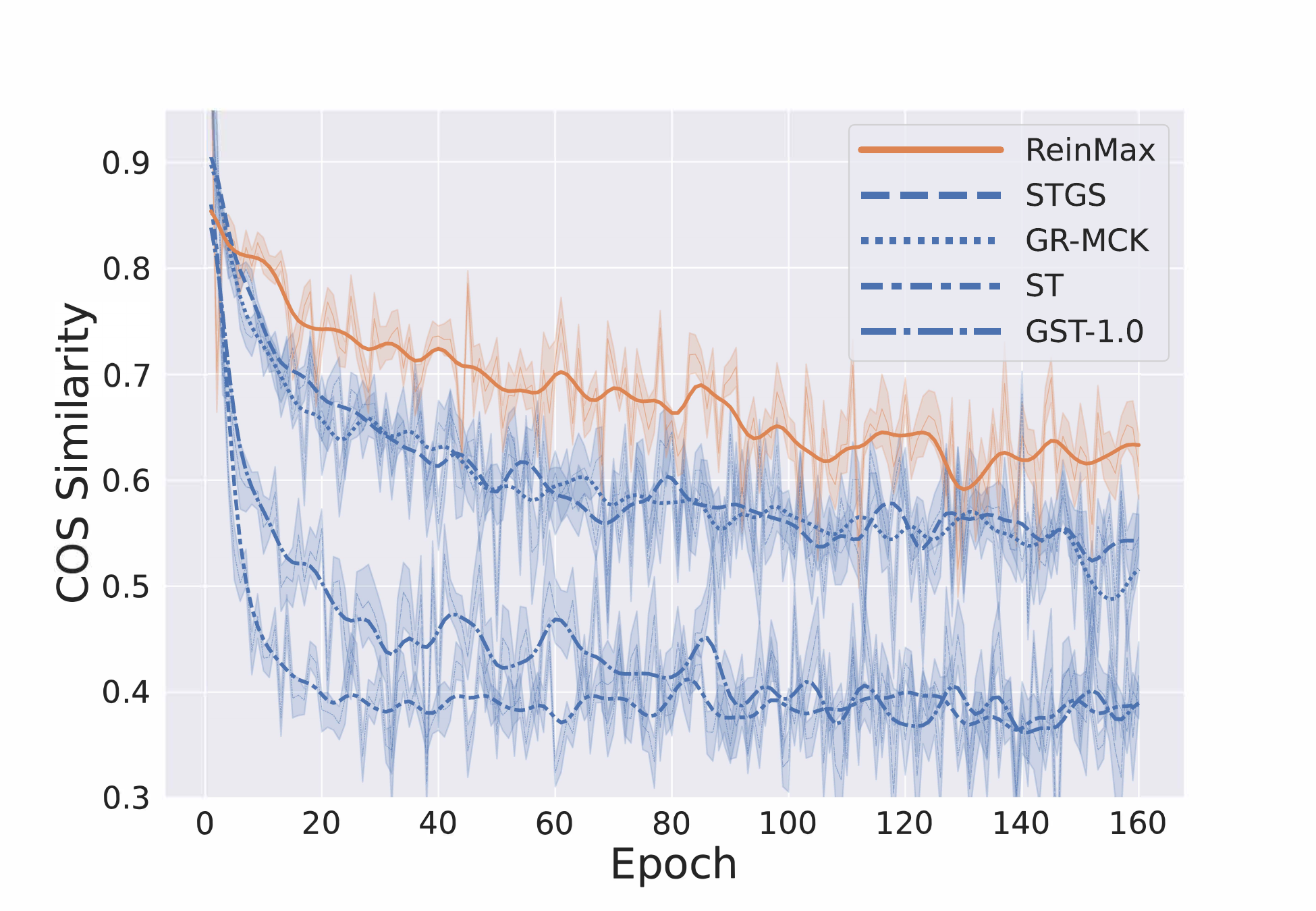}
        \vspace{-4mm}
        \label{}
    \end{subfigure}
    \vspace{-2mm}
    \caption{
    The training $-$ELBO (left) and the cos similarity between the gradient and its approximations (right) on MNIST-VAE (with 4 latent dimensions and 8 categorical dimensions).  
    }
    \vspace{-3mm}
    \label{fig:mnist-8-4}
\end{figure}

\subsection{ListOps}
We conducted unsupervised parsing on ListOps~\citep{Nangia2018ListOpsAD} and summarized the average accuracy and the standard derivation in Table~\ref{table:listops}. 
We also visualized the accuracy and loss on the valid set in Figure~\ref{fig:listops}. 
Although the ST algorithm performs poorly on polynomial programming, it achieves a reasonable performance on this task. 
Also, while all baseline methods perform similarly, our proposed method stands out and brings consistent improvements. 
This further demonstrates the benefits of achieving second-order accuracy and the effectiveness of our proposed method.

\begin{table}[t]

\centering
\caption{Average time cost (per epoch) / peak memory consumption on quadratic programming (QP) and MNIST-VAE. QP is configured to have 128 binary latent variables and 512 samples per batch. MNIST-VAE is configured to have 10 categorical dimensions and 30 latent dimensions.}
\label{table:efficiency}
\scalebox{.73}{
\begin{tabular}{lccccccc}
\toprule
  & \ours & ST & STGS & GST-1.0 & GR-MCK$_{100}$ & GR-MCK$_{300}$ & GR-MCK$_{1000}$ \\ 
  \midrule
QP & 0.2s / 6.5Mb & 0.2s / 5.0Mb & 0.2s / 5.5Mb & 0.2s / 8.0Mb & 0.8s / 0.3Gb & 2.2s / 1Gb & 6.6s / 3Gb\\
  \midrule
{\small MNIST-VAE} & 5.2s / 13Mb & 5.2s / 13Mb & 5.2s / 13Mb & 5.2s / 13Mb & 5.2s / 76Mb & 5.2s / 0.2Gb & 5.4s / 0.6Gb \\
\bottomrule
\end{tabular}
}
\vspace{-.3cm}
\end{table}

\subsection{MNIST-VAE}
We benchmark the performance by training variational auto-encoders (VAE) with \emph{categorical} latent variables on MNIST~\citep{LeCun1998GradientbasedLA}. 
As we aim to compare gradient estimators, we focus our discussions on training ELBO.
We find that training performance largely mirrors test performance~\citep{Dong2020DisARMAA,Dong2021CoupledGE,Fan2022TrainingDD} and briefly discussed test ELBO in Appendix~\ref{appendix:exp}. 

\smallsection{Biases of the Approximated Gradient}
With 4 latent dimensions and 8 categorical dimensions, 
we iterate through the whole latent space (the size of the latent space is only $4096$), compute the gradient as in Equation~\ref{eqn:exact}, 
and measured the cosine similarity between the gradient of latent variables and various approximations.
As visualized in Figure~\ref{fig:mnist-8-4}, \ours achieves consistently more accurate gradient approximation across the training and, accordingly, faster convergence. 
Also, we can observe that, besides faster convergence, the performance of \ours is more stable. 

\smallsection{Experiment with Larger Latent Spaces}
Let us proceed to larger latent spaces.
First, we consider 4 settings with the latent space of $2^{48}$. 
Then, following~\citet{Fan2022TrainingDD}, we also conduct experiments with 10 latent dimensions and 30 categorical dimensions (the size of the latent space is $10^{30}$). 
As summarized in Table~\ref{table:mnist-vae}, \ours achieves the best performance on all configurations. 

\smallsection{GST-1.0 Performance on Different Problems}
It is worth mentioning that, despite GST-1.0 achieving good performance on most settings of MNIST-VAE, it fails to maintain this performance on polynomial programming and unsupervised parsing, as discussed before. 
Upon discussing with \citet{Fan2022TrainingDD}, we suggest that this phenomenon is caused by the characteristic of GST-1.0, which behaves similarly to ST on problems with a near one-hot optimal distribution. 
In other words, GST-1.0 has an implicit prior and prefers distributions that are not one-hot. 
At the same time, a different variant of GST (i.e., GST-p) would behave similarly to STGS on problems with a near one-hot optimal distribution, which achieves a significant performance boost over GST-1.0 on polynomial programming. 
However, on MNIST-VAE and ListOps, GST-p achieves an inferior performance. 

This observation verifies our intuition that, without understanding the mechanism of ST, different applications have different preferences on its configurations.
Meanwhile, ReinMax achieves consistent improvements in all settings, which greatly simplifies future algorithms developments.

\begin{table}[t!]

\centering
\caption{Performance on NATS-Bench. $^*$ Baseline results are referenced from \citet{Dong2020NATSBenchBN}.}
\label{table:nas}
\scalebox{0.85}{
\begin{tabular}{l|cc|cc|cc}
\toprule
  & \multicolumn{2}{c|}{CIFAR-10} & \multicolumn{2}{c|}{CIFAR-100} & \multicolumn{2}{c}{ImageNet-16-120} \\ 
  \cmidrule{2-7}
  & validation & test & validation & test & validation & test \\ 
\midrule
GDAS + STGS$^*$ & 89.68$\pm$0.72 & 93.23$\pm$0.58 & 68.35$\pm$2.71 & 68.17$\pm$2.50 & 39.55$\pm$0.00 & 39.40$\pm$0.00 \\
\midrule
GDAS + \ours & \textbf{90.01$\pm$0.12} & \textbf{93.44$\pm$0.23} & \textbf{69.29$\pm$2.34} & \textbf{69.41$\pm$2.24} & \textbf{41.47$\pm$0.79} & \textbf{42.03$\pm$0.41} \\
\bottomrule
\end{tabular}
}
\end{table}

\begin{table}[t!]

\centering
\caption{Training $-$ELBO on MNIST. $^*$ All baseline results are referenced from \citet{Fan2022TrainingDD}}
\label{table:mnist-vae-32x64}
\scalebox{0.75}{
\begin{tabular}{lcccccccc}
\toprule
  & RLOO$^*$ & DisARM-Tree$^*$ & STGS$^*$ & GR-MCK$^*$ & GST-1.0$^*$ & ST$^*$ & \ours \\ 
\midrule
Neg. ELBO & 104.03$\pm$0.23 & 103.10$\pm$0.25 & 97.32$\pm$0.20 & 110.74$\pm$1.23 & 96.09$\pm$0.25 & 116$\pm$0.09 & \textbf{93.44$\pm$0.51} \\
\bottomrule
\end{tabular}
}
\vspace{-0.3cm}
\end{table}

\subsection{Applying \ours to Differentiable Neural Architecture Search}
To demonstrate the applicability of ReinMax as a drop-in replacement, we conduct experiments following the topology search setting in the NATS-Bench benchmark~\citep{Dong2020NATSBenchBN}, and summarize the results in Table~\ref{table:nas}. 
GDAS is an algorithm that employs STGS to estimate the gradient of neural architecture parameters~\citep{Dong2019SearchingFA}.
We replaced STGS with ReinMax as the gradient estimator (configurations elaborated in Appendix~\ref{appendix:exp}).
ReinMax brings consistent performance improvements across all three datasets, demonstrating the great potential of ReinMax.

\begin{figure}[t!]
    \centering
    \begin{subfigure}[t]{0.32\linewidth}
        \centering
        \includegraphics[width=1.0\textwidth]{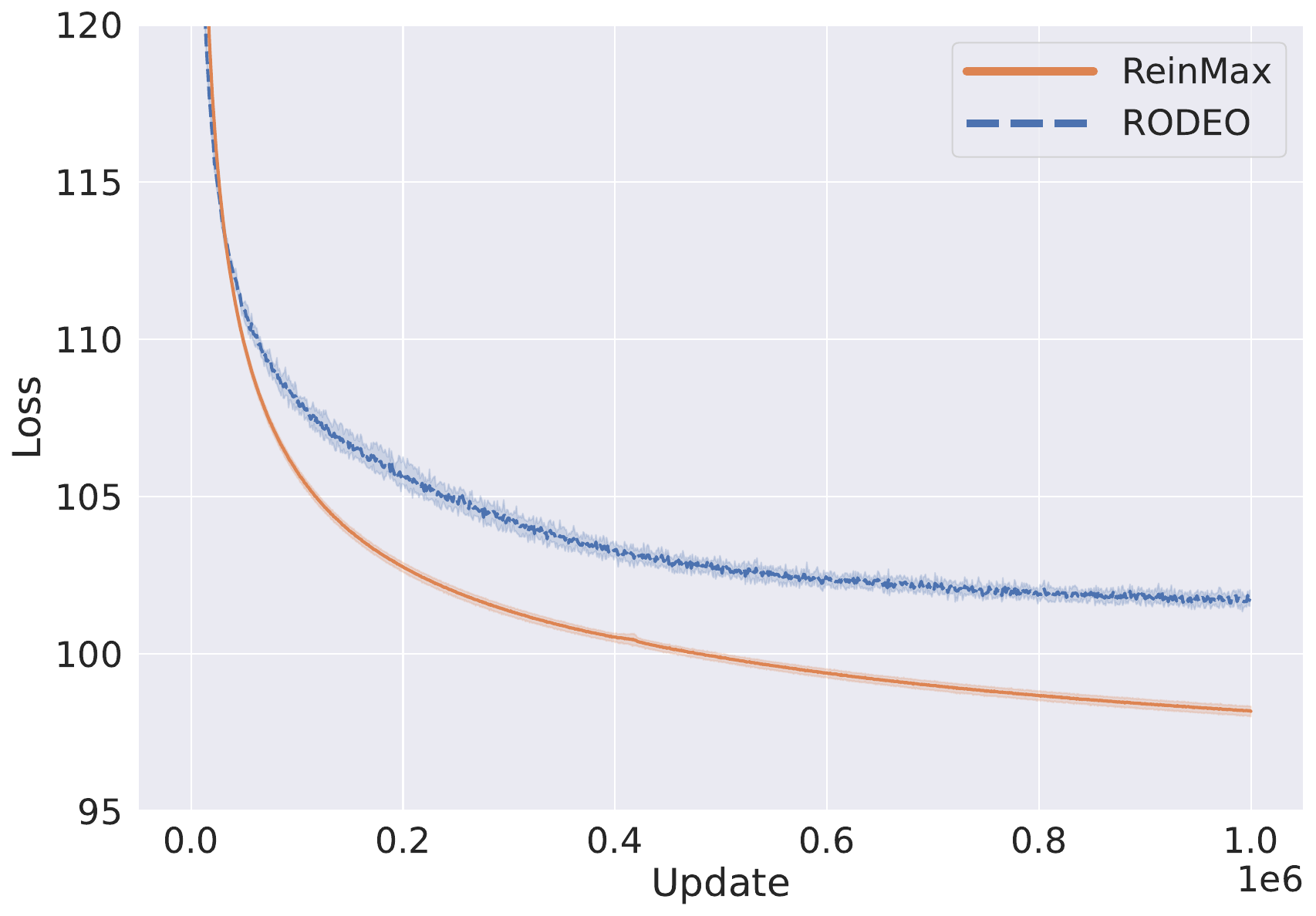}
        \vspace{-4mm}
        \caption{MNIST when K=2.}
        \label{}
    \end{subfigure}
    \hfill
    \begin{subfigure}[t]{0.32\linewidth}
        \centering
        \includegraphics[width=1.0\textwidth]{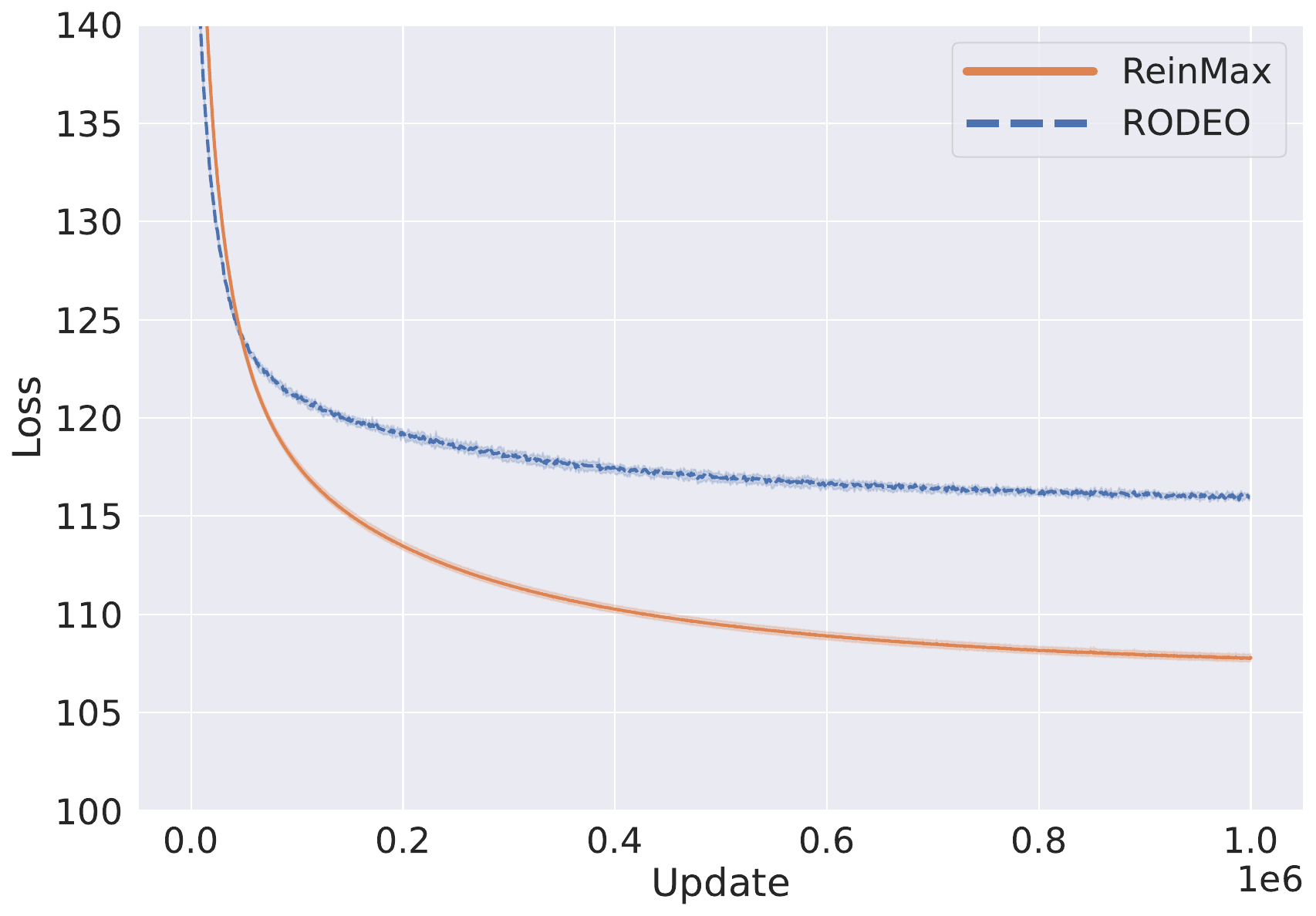}
        \vspace{-4mm}
        \caption{Omniglot when K=2.}
        \label{}
    \end{subfigure}
    \hfill
    \begin{subfigure}[t]{0.32\linewidth}
        \centering
        \includegraphics[width=1.0\textwidth]{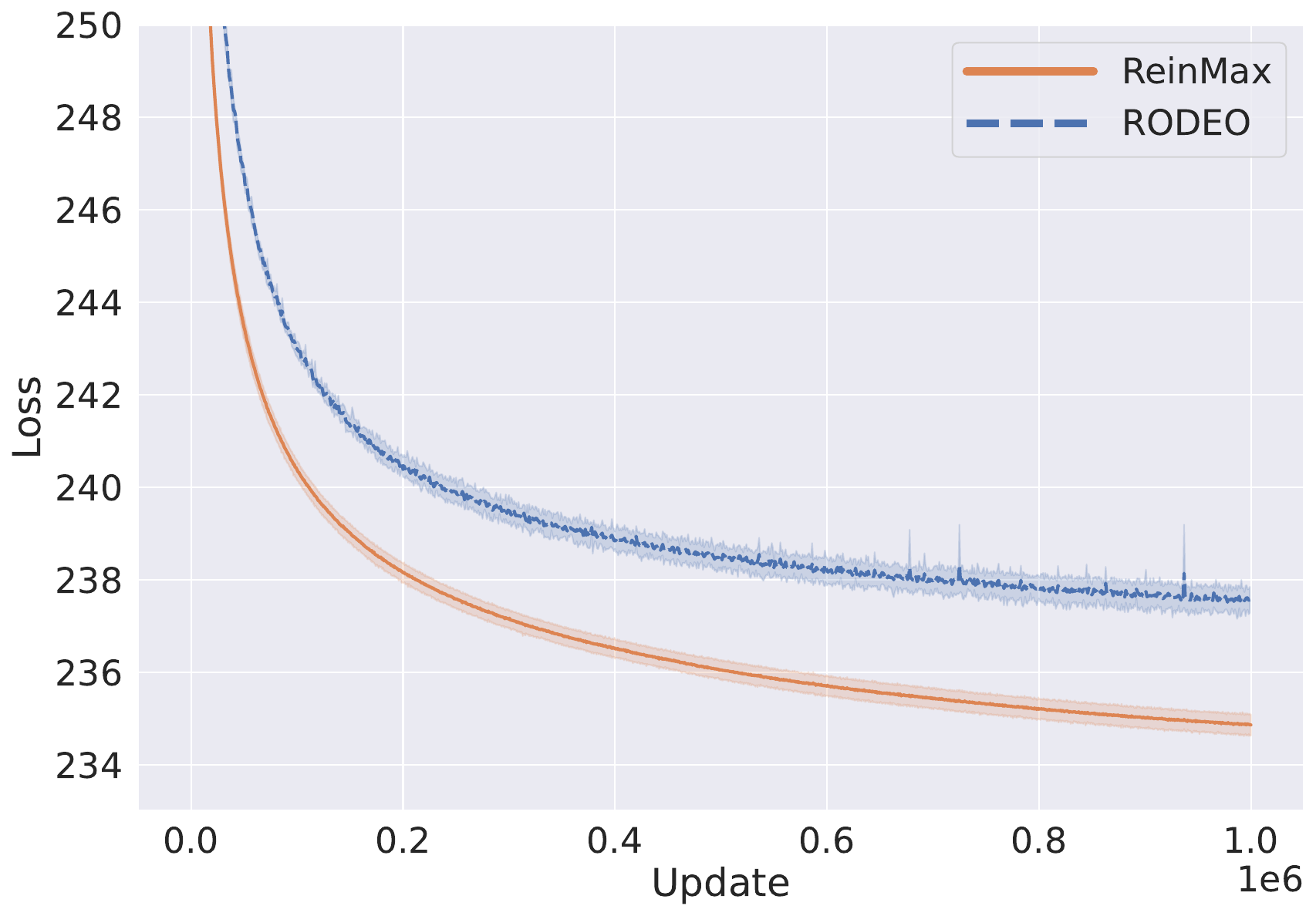} 
        \vspace{-4mm}
        \caption{Fashion-MNIST when K=2.}
        \label{}
    \end{subfigure}
    \begin{subfigure}[t]{0.32\linewidth}
        \centering
        \includegraphics[width=1.0\textwidth]{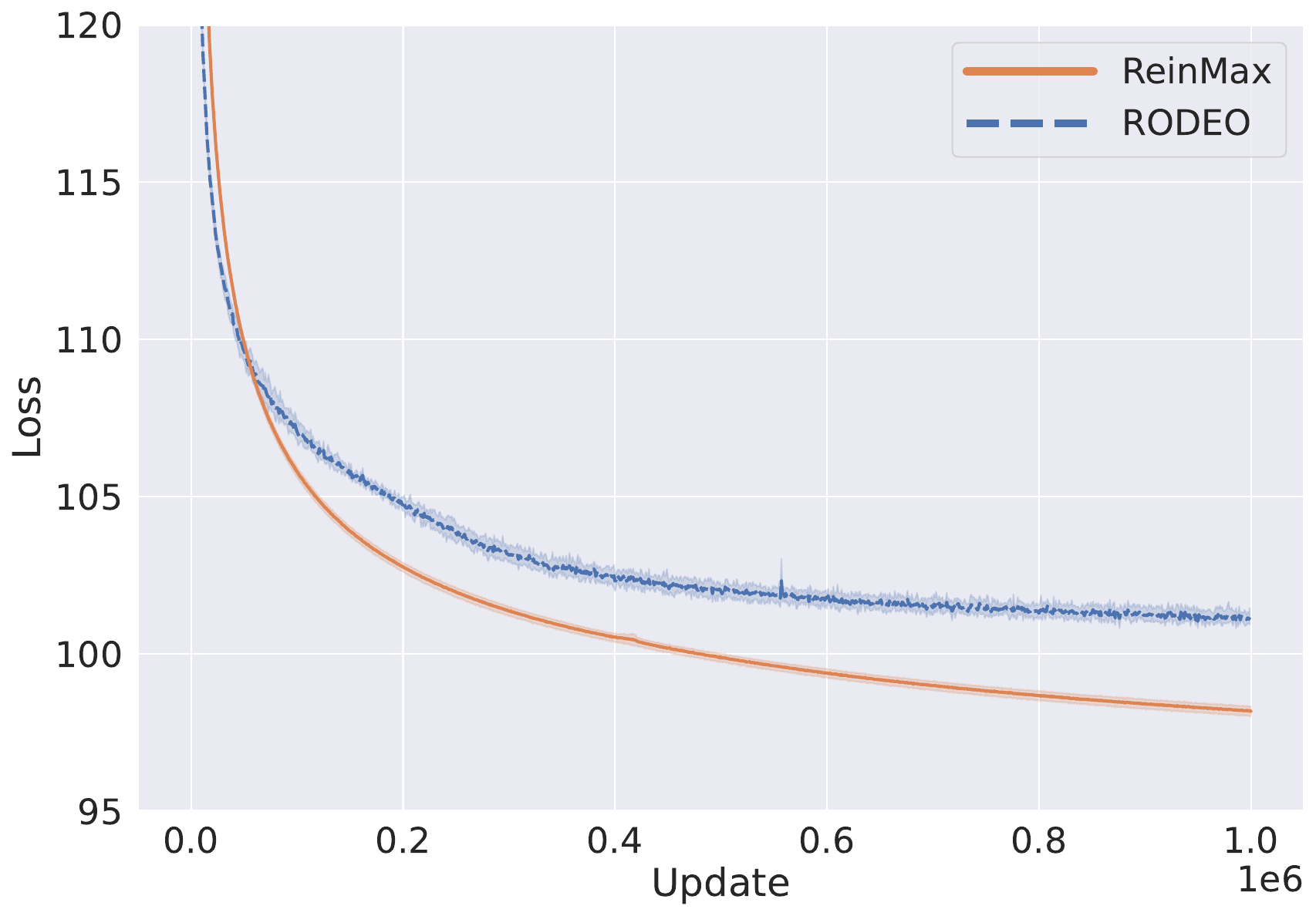}
        \vspace{-4mm}
        \caption{MNIST when K=3.}
        \label{}
    \end{subfigure}
    \hfill
    \begin{subfigure}[t]{0.32\linewidth}
        \centering
        \includegraphics[width=1.0\textwidth]{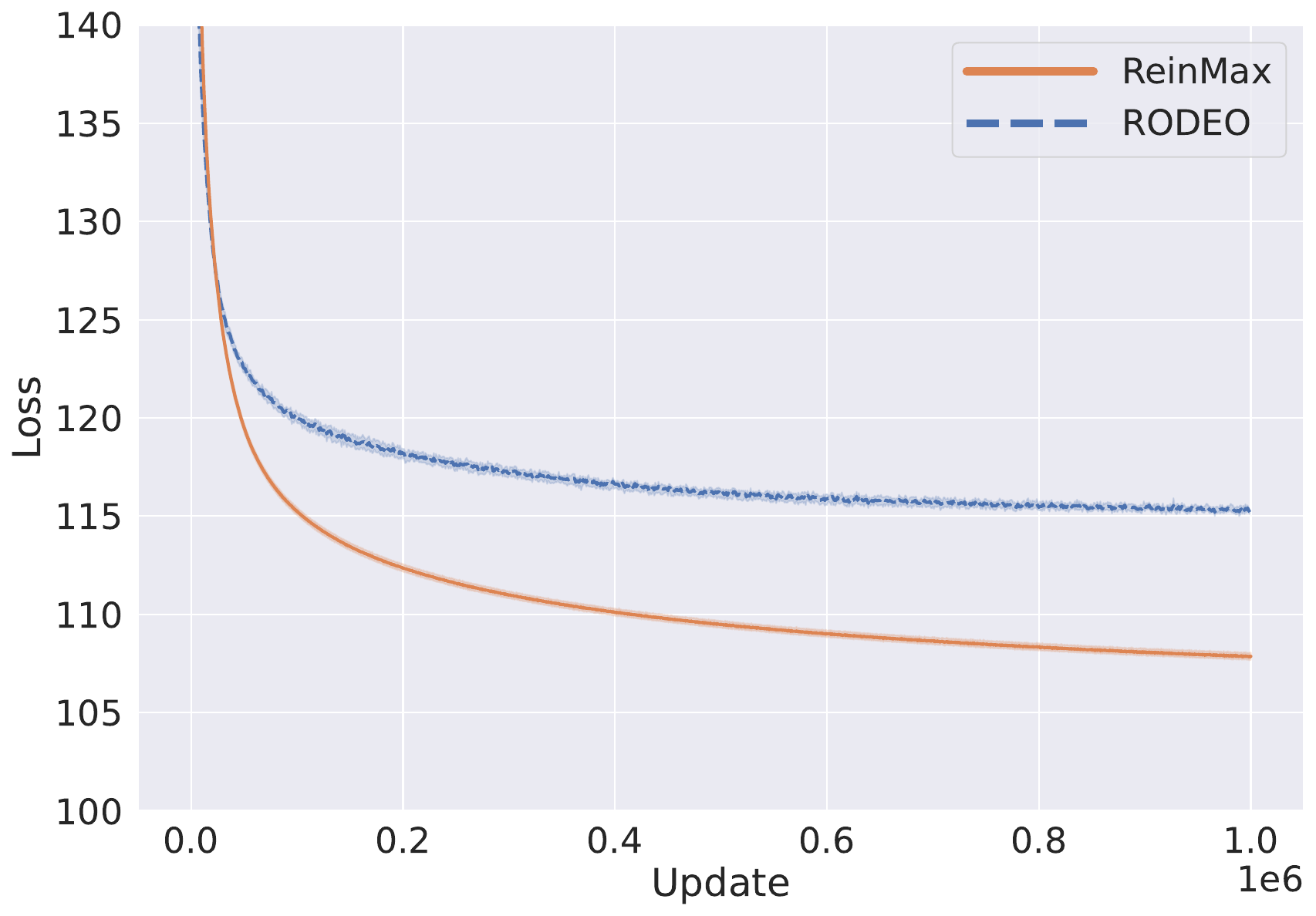}
        \vspace{-4mm}
        \caption{Omniglot when K=3.}
        \label{}
    \end{subfigure}
    \hfill
    \begin{subfigure}[t]{0.32\linewidth}
        \centering
        \includegraphics[width=1.0\textwidth]{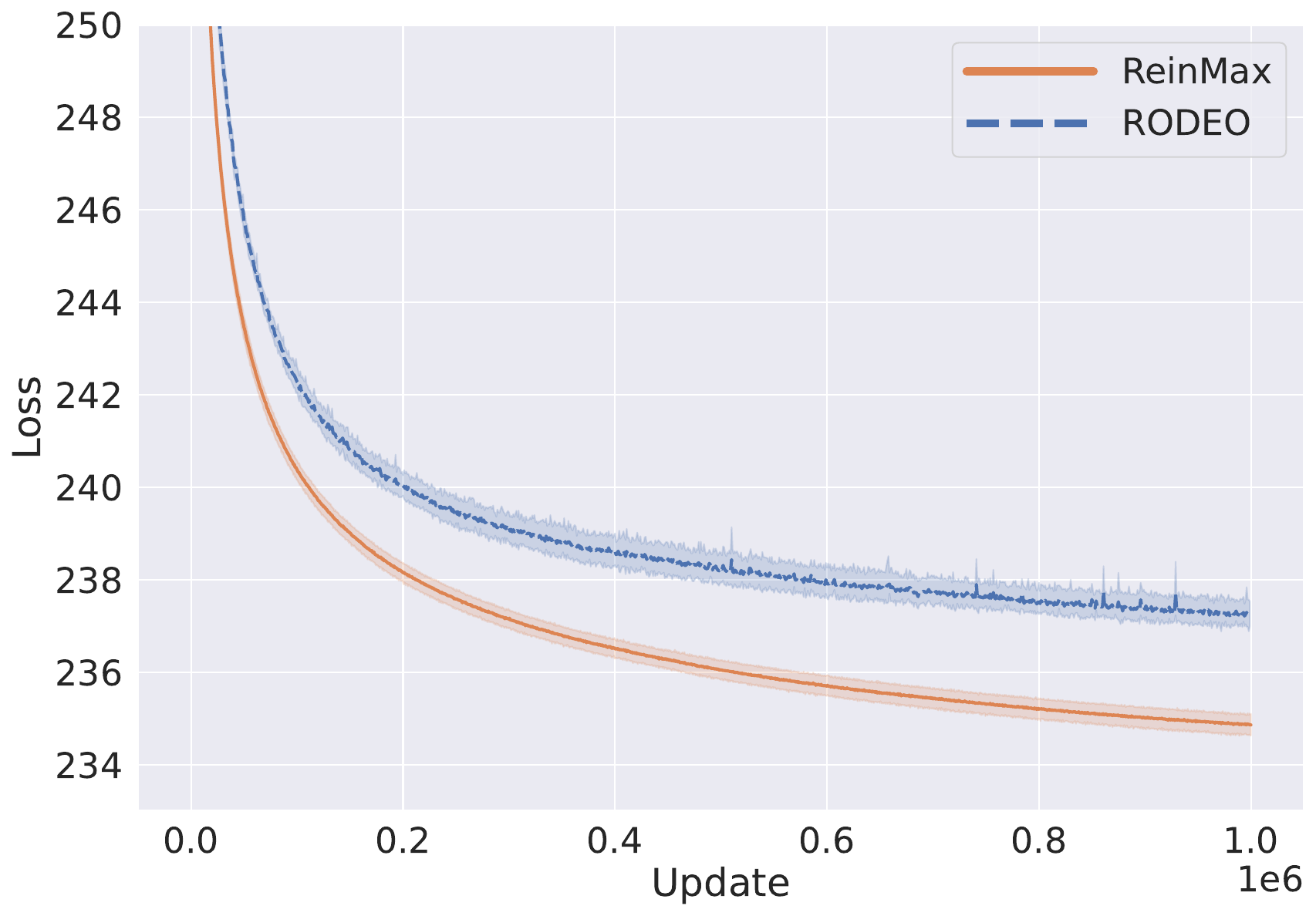} 
        \vspace{-4mm}
        \caption{Fashion-MNIST when K=3.}
        \label{}
    \end{subfigure}
    \caption{
    2$\times$200 VAE training curves on MNIST, Omniglot, and Fashion-MNIST when K=2 or 3. 
    }
    \label{fig:vae-rodeo}
\end{figure}

\begin{table}[t!]

\centering
\caption{Train $-$ELBO of $2\times200$ VAE on MNIST, Fashion-MNIST, and Omniglot. $^*$ Baseline results are referenced from ~\cite{shi2022gradient}. K refers to the number of evaluations. }
\label{table:rodeo}
\scalebox{.8}{
\begin{tabular}{l|l|cccc cc}
\toprule
~ & ~ & RELAX$^*$ & ARMS$^*$ & DisARM$^*$ & Double CV$^*$ & RODEO$^*$ & ReinMax \\
\midrule
\multirow{3}{*}{K=3} & MNIST & 101.99$\pm$0.04 & 100.84$\pm$0.14 & / & 100.94$\pm$0.09 & 100.46$\pm$0.13 & \textbf{97.83$\pm$0.36} \\
& Fashion-MNIST & 237.74$\pm$0.12 & 237.05$\pm$0.12 & / & 237.40$\pm$0.11 & 236.88$\pm$0.12 & \textbf{234.53$\pm$0.42} \\
& Omniglot & 115.70$\pm$0.08 & 115.32$\pm$0.07 & / & 115.06$\pm$0.12 & 115.01$\pm$0.05 & \textbf{107.51$\pm$0.42} \\
\midrule
\multirow{3}{*}{K=2} & MNIST & / & / & 102.75$\pm$0.08 & 102.14$\pm$0.06 & 101.89$\pm$0.17 & \textbf{98.17$\pm$0.29} \\
& Fashion-MNIST & / & / & 237.68$\pm$0.13 & 237.55$\pm$0.16 & 237.44$\pm$0.09 & \textbf{234.89$\pm$0.21} \\
& Omniglot & / & / & 116.50$\pm$0.04 & 116.39$\pm$0.10 & 115.93$\pm$0.06 & \textbf{107.79$\pm$0.27}\\
\bottomrule
\end{tabular}
}
\vspace{-0.2cm}
\end{table}

\subsection{Comparisons with REINFORCE-style Methods}
\label{subsec:exp_reinforce}

Here, we conduct experiments to discuss the difference between \ours and REINFORCE-style methods. 
First, following ~\citet{Fan2022TrainingDD}, we conduct experiments on the setting with a larger batch size (i.e., 200), longer training (i.e., $5\times 10^5$ steps), 32 latent dimensions, and 64 categorical dimensions (details are elaborated in Appendix~\ref{appendix:exp}). 
As in Table~\ref{table:mnist-vae-32x64}, \ours outperforms all baselines, including two REINFORCE-based methods~\citep{Dong2020DisARMAA, Dong2021CoupledGE}.

We further conduct experiments to compare with the state of the art. 
Specifically we apply ReinMax to Bernoulli VAEs on MNIST, Fashion-MNIST~\citep{Xiao2017FashionMNISTAN}, and Omniglot\citep{Lake2015HumanlevelCL}, adhering closely to the experimental settings of \citet{shi2022gradient}, including pre-processing, model architecture, batch size, and training epochs. 
As in Tables~\ref{table:rodeo} and Figure~\ref{fig:vae-rodeo}, ReinMax consistently outperforms RODEO across all settings.
To better understand the difference between RODEO and ReinMax, we conduct more experiments on polynomial programming (as elaborated in Appendix~\ref{appendix:rodeo}). 

Overall, ReinMax achieves better performance in more challenging scenarios, i.e., smaller batch size, more latent variables, or more complicated problems. Meanwhile, REINFORCE and RODEO achieve better performance on simpler problem settings, i.e., larger batch size, fewer latent variables, or simpler problems. This observation matches our intuition:
\begin{itemize}[leftmargin=*]
    \item
    \vspace{-0.2cm}
    REIFORCE-style algorithms excel as they provide unbiased gradient estimation but may fall short in complex scenarios, since they only utilize the zero-order information (i.e., a scalar $f(\mD)$).
    
    \item 
    ReinMax, using more information (i.e., a vector $\frac{\partial f(\mD)}{\partial \mD}$), handles challenging scenarios better. Meanwhile, 
    as a consequence of its estimation bias, ReinMax leads to slower convergence in some simple scenarios.
\end{itemize}

\subsection{Discussions}
\label{subsec: exp-discussions}
\begin{wrapfigure}{r}{0.42 \textwidth} 
\centering
\vspace{-0.9cm}
\includegraphics[width=0.37 \textwidth]{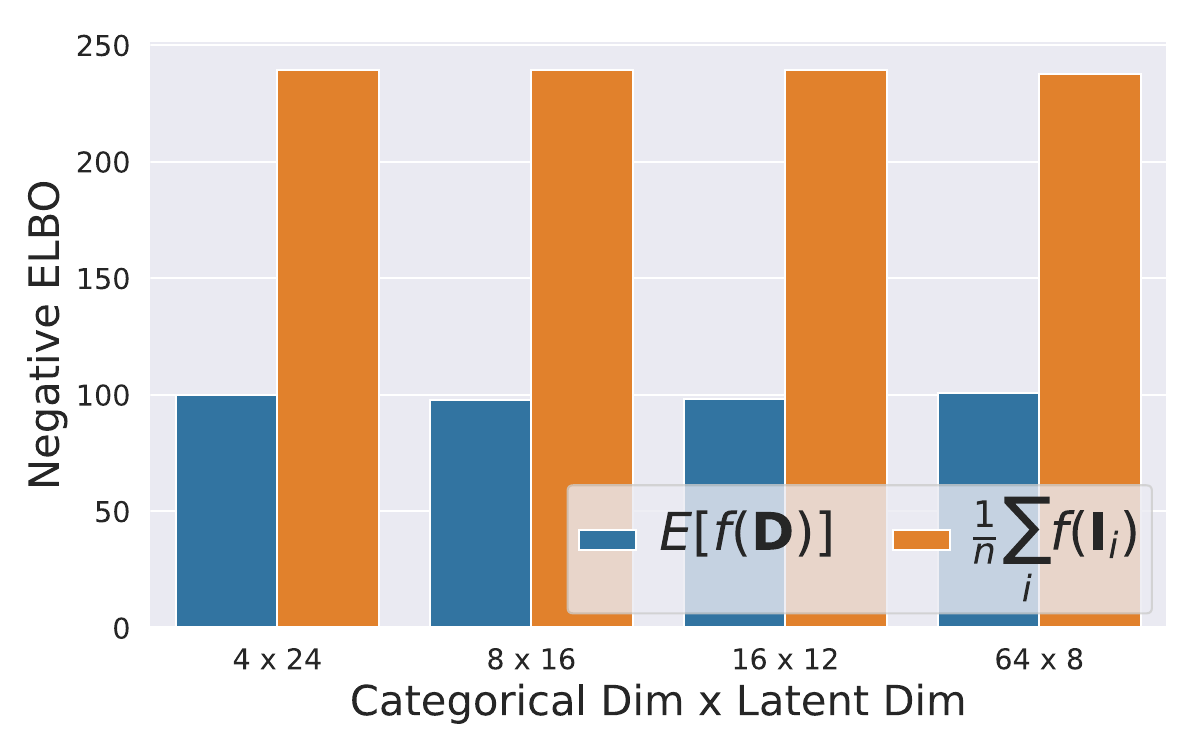}
\vspace{-0.2cm}
\caption{ Training $-$ELBO on MNIST-VAE when using $\frac{1}{n} \sum_i f(\mI_i)$ and $E[f(\mD)]$ as baselines respectively. }
\vspace{-0.8cm}
\label{fig:baseline}
\end{wrapfigure}

\smallsection{Choice of Baseline}
As introduced in Section~\ref{subsec: altervative-first-order}, the choice of subtraction baseline has a huge impact on the performance. 
Here, we demonstrate this empirically. 

We use $\frac{1}{n} \sum_i f(\mI_i)$ as the baseline and compare the resulting gradient approximation with ReinMax.
As visualized  in Figure~\ref{fig:baseline}, ReinMax, which uses $E[f(\mD)]$ as the baseline, significantly outperforms the one that uses $\frac{1}{n} \sum_i f(\mI_i)$ as the baseline. 
We suspect that the gradient approximation using $\frac{1}{n} \sum_i f(\mI_i)$ as the baseline is very unstable as it contains the $\frac{1}{n \cdot p(\mD)}$ term.

\smallsection{Temperature Scaling}
On MNIST-VAE (four settings with the $2^{48}$ latent space), we utilize heatmaps to visualize the final performance of all five methods under different temperatures, i.e., $\{0.1, 0.3, 0.5, 0.7, 1, 2, 3, 4, 5\}$. 
As in Figure~\ref{fig:temperature}, these methods have different preferences for the temperature configuration.
Specifically, STGS, GST-1.0, and GR-MCK prefer to set the temperature $\tau \leq 1$.
Differently, ST and \ours prefer to set the temperature $\tau \geq 1$. 
These observations match our analyses in Section~\ref{sec:temperature} that a small $\tau$ can help reduce the bias introduced by STGS-style methods. 
Also, it verifies that ST and \ours work differently from STGS, GST-1.0, and GR-MCK. 

\smallsection{Efficiency}
As summarized in Table~\ref{table:efficiency}, 
we can observe that, 
since GR-MCK uses the Monte Carlo method to reduce the variance, it has larger time and memory consumption, which becomes less significant with fewer Monte Carlo samples (we use GR-MCK$_{s}$ to indicate GR-MCK with $s$ Monte Carlo samples). 
Meanwhile, all remaining methods have roughly the same time and memory consumption. 
This shows that the computation overheads of ReinMax are negligible. 

\section{Conclusion and Future Work}
\label{sect:conclusion}

In this study, we seek the underlying principle of the Straight-Through (ST) gradient estimator. 
We formally establish that ST works as a first-order approximation of the gradient and propose a novel method, ReinMax, which incorporates Heun's Method and achieves second-order accuracy without requiring second-order derivatives. 
We conduct extensive experiments on polynomial programming, unsupervised generative modeling, and structured output prediction.
\ours brings consistent improvements over the state-of-the-art methods. 

It is worth mentioning that analyses in this study further guided us to empower Mixture-of-Expert training~\citep{Liu2023SparseMixer}. 
Specifically, for gradient approximation of sparse expert routing, while \ours requires the network to be fully activated, \citet{Liu2023SparseMixer} uses $f(\mZero)$ as the baseline and only requires the network to be partially activated. 
In the future, we plan to conduct further analyses on the truncation error to stabilize and improve the gradient estimation. 
\section*{Acknowledgement}
We would like to thank all reviewers for their constructive comments, the engineering team at Microsoft for providing computation infrastructure support, 
Alessandro Sordoni, Nicolas Le Roux, and Greg Yang for their helpful discussions. 

\bibliography{icml2022}

\begin{thebibliography}{39}
\providecommand{\natexlab}[1]{#1}
\providecommand{\url}[1]{\texttt{#1}}
\expandafter\ifx\csname urlstyle\endcsname\relax
  \providecommand{\doi}[1]{doi: #1}\else
  \providecommand{\doi}{doi: \begingroup \urlstyle{rm}\Url}\fi

\bibitem[Ascher \& Petzold(1998)Ascher and Petzold]{Ascher1998ComputerMF}
Ascher, U.~M. and Petzold, L.~R.
\newblock Computer methods for ordinary differential equations and
  differential-algebraic equations.
\newblock 1998.

\bibitem[Bengio et~al.(2013)Bengio, L{\'e}onard, and
  Courville]{Bengio2013EstimatingOP}
Bengio, Y., L{\'e}onard, N., and Courville, A.~C.
\newblock Estimating or propagating gradients through stochastic neurons for
  conditional computation.
\newblock \emph{ArXiv}, abs/1308.3432, 2013.

\bibitem[Choi et~al.(2017)Choi, Yoo, and goo Lee]{Choi2017LearningTC}
Choi, J., Yoo, K.~M., and goo Lee, S.
\newblock Learning to compose task-specific tree structures.
\newblock In \emph{AAAI}, 2017.

\bibitem[Chung et~al.(2017)Chung, Ahn, and Bengio]{Chung2016HierarchicalMR}
Chung, J., Ahn, S., and Bengio, Y.
\newblock Hierarchical multiscale recurrent neural networks.
\newblock In \emph{ICLR}, 2017.

\bibitem[Dayan et~al.(1995)Dayan, Hinton, Neal, and Zemel]{Dayan1995TheHM}
Dayan, P., Hinton, G.~E., Neal, R.~M., and Zemel, R.~S.
\newblock The helmholtz machine.
\newblock \emph{Neural Computation}, 7:\penalty0 889--904, 1995.

\bibitem[Dong \& Yang(2019)Dong and Yang]{Dong2019SearchingFA}
Dong, X. and Yang, Y.
\newblock Searching for a robust neural architecture in four gpu hours.
\newblock \emph{CVPR}, 2019.

\bibitem[Dong et~al.(2020{\natexlab{a}})Dong, Liu, Musial, and
  Gabrys]{Dong2020NATSBenchBN}
Dong, X., Liu, L., Musial, K., and Gabrys, B.
\newblock Nats-bench: Benchmarking nas algorithms for architecture topology and
  size.
\newblock \emph{TPAMI}, 2020{\natexlab{a}}.

\bibitem[Dong et~al.(2020{\natexlab{b}})Dong, Mnih, and
  Tucker]{Dong2020DisARMAA}
Dong, Z., Mnih, A., and Tucker, G.
\newblock Disarm: An antithetic gradient estimator for binary latent variables.
\newblock In \emph{NeurIPS}, 2020{\natexlab{b}}.

\bibitem[Dong et~al.(2021)Dong, Mnih, and Tucker]{Dong2021CoupledGE}
Dong, Z., Mnih, A., and Tucker, G.
\newblock Coupled gradient estimators for discrete latent variables.
\newblock In \emph{NeurIPS}, 2021.

\bibitem[Fan et~al.(2022)Fan, Chi, Rudnicky, and Ramadge]{Fan2022TrainingDD}
Fan, T.-H., Chi, T.-C., Rudnicky, A.~I., and Ramadge, P.~J.
\newblock Training discrete deep generative models via gapped straight-through
  estimator.
\newblock In \emph{ICML}, 2022.

\bibitem[Fedus et~al.(2021)Fedus, Zoph, and Shazeer]{Fedus2021SwitchTS}
Fedus, W., Zoph, B., and Shazeer, N.~M.
\newblock Switch transformers: Scaling to trillion parameter models with simple
  and efficient sparsity.
\newblock \emph{ArXiv}, abs/2101.03961, 2021.

\bibitem[Grathwohl et~al.(2018)Grathwohl, Choi, Wu, Roeder, and
  Duvenaud]{Grathwohl2017BackpropagationTT}
Grathwohl, W., Choi, D., Wu, Y., Roeder, G., and Duvenaud, D.~K.
\newblock Backpropagation through the void: Optimizing control variates for
  black-box gradient estimation.
\newblock In \emph{ICLR}, 2018.

\bibitem[Gregor et~al.(2014)Gregor, Danihelka, Mnih, Blundell, and
  Wierstra]{Gregor2013DeepAN}
Gregor, K., Danihelka, I., Mnih, A., Blundell, C., and Wierstra, D.
\newblock Deep autoregressive networks.
\newblock In \emph{ICML}, 2014.

\bibitem[Gu et~al.(2016)Gu, Levine, Sutskever, and Mnih]{Gu2015MuPropUB}
Gu, S.~S., Levine, S., Sutskever, I., and Mnih, A.
\newblock Muprop: Unbiased backpropagation for stochastic neural networks.
\newblock In \emph{ICLR}, 2016.

\bibitem[Gumbel(1954)]{Gumbel1954StatisticalTO}
Gumbel, E.~J.
\newblock Statistical theory of extreme values and some practical applications
  : A series of lectures.
\newblock 1954.

\bibitem[Jang et~al.(2017)Jang, Gu, and Poole]{Jang2016CategoricalRW}
Jang, E., Gu, S.~S., and Poole, B.
\newblock Categorical reparameterization with gumbel-softmax.
\newblock In \emph{ICLR}, 2017.

\bibitem[Kingma \& Ba(2015)Kingma and Ba]{Kingma2014AdamAM}
Kingma, D.~P. and Ba, J.
\newblock Adam: A method for stochastic optimization.
\newblock In \emph{ICLR}, 2015.

\bibitem[Kingma \& Welling(2013)Kingma and Welling]{Kingma2013AutoEncodingVB}
Kingma, D.~P. and Welling, M.
\newblock Auto-encoding variational bayes.
\newblock \emph{CoRR}, abs/1312.6114, 2013.

\bibitem[Lake et~al.(2015)Lake, Salakhutdinov, and
  Tenenbaum]{Lake2015HumanlevelCL}
Lake, B.~M., Salakhutdinov, R., and Tenenbaum, J.~B.
\newblock Human-level concept learning through probabilistic program induction.
\newblock \emph{Science}, 2015.

\bibitem[LeCun et~al.(1998)LeCun, Bottou, Bengio, and
  Haffner]{LeCun1998GradientbasedLA}
LeCun, Y., Bottou, L., Bengio, Y., and Haffner, P.
\newblock Gradient-based learning applied to document recognition.
\newblock \emph{Proc. IEEE}, 1998.

\bibitem[Liu et~al.(2019)Liu, Simonyan, and Yang]{liu2018darts}
Liu, H., Simonyan, K., and Yang, Y.
\newblock Darts: Differentiable architecture search.
\newblock In \emph{ICLR}, 2019.

\bibitem[Liu et~al.(2020)Liu, Jiang, He, Chen, Liu, Gao, and Han]{Liu2019OnTV}
Liu, L., Jiang, H., He, P., Chen, W., Liu, X., Gao, J., and Han, J.
\newblock On the variance of the adaptive learning rate and beyond.
\newblock In \emph{ICLR}, 2020.

\bibitem[Liu et~al.(2023)Liu, Gao, and Chen]{Liu2023SparseMixer}
Liu, L., Gao, J., and Chen, W.
\newblock Sparse backpropagation for moe training.
\newblock \emph{ArXiv}, abs/2310.00811, 2023.

\bibitem[Maddison et~al.(2014)Maddison, Tarlow, and Minka]{Maddison2014AS}
Maddison, C.~J., Tarlow, D., and Minka, T.~P.
\newblock A* sampling.
\newblock In \emph{NIPS}, 2014.

\bibitem[Mullin \& Rosenblatt(1962)Mullin and
  Rosenblatt]{Mullin1962PrinciplesON}
Mullin, A.~A. and Rosenblatt, F.
\newblock Principles of neurodynamics.
\newblock 1962.

\bibitem[Nangia \& Bowman(2018)Nangia and Bowman]{Nangia2018ListOpsAD}
Nangia, N. and Bowman, S.~R.
\newblock Listops: A diagnostic dataset for latent tree learning.
\newblock \emph{ArXiv}, abs/1804.06028, 2018.

\bibitem[Neal(1992)]{Neal1992ConnectionistLO}
Neal, R.~M.
\newblock Connectionist learning of belief networks.
\newblock \emph{Artif. Intell.}, 56:\penalty0 71--113, 1992.

\bibitem[Paulus et~al.(2021)Paulus, Maddison, and
  Krause]{Paulus2020RaoBlackwellizingTS}
Paulus, M.~B., Maddison, C.~J., and Krause, A.
\newblock Rao-blackwellizing the straight-through gumbel-softmax gradient
  estimator.
\newblock In \emph{ICLR}, 2021.

\bibitem[Pervez et~al.(2020)Pervez, Cohen, and Gavves]{Pervez2020LowBL}
Pervez, A., Cohen, T., and Gavves, E.
\newblock Low bias low variance gradient estimates for boolean stochastic
  networks.
\newblock In \emph{ICML}, 2020.

\bibitem[Rennie et~al.(2016)Rennie, Marcheret, Mroueh, Ross, and
  Goel]{Rennie2016SelfCriticalST}
Rennie, S.~J., Marcheret, E., Mroueh, Y., Ross, J., and Goel, V.
\newblock Self-critical sequence training for image captioning.
\newblock In \emph{CVPR}, 2016.

\bibitem[Rosenblatt(1957)]{rosenblatt1957perceptron}
Rosenblatt, F.
\newblock \emph{The perceptron, a perceiving and recognizing automaton Project
  Para}.
\newblock Cornell Aeronautical Laboratory, 1957.

\bibitem[Rumelhari et~al.(1986)Rumelhari, Hintont, Ronald, J., and
  Williams]{Rumelhari2004LearningRB}
Rumelhari, D.~E., Hintont, G.~E., Ronald, J., and Williams.
\newblock Learning representations by backpropagating errors.
\newblock \emph{Nature}, 323:\penalty0 533–536, 1986.

\bibitem[Shi et~al.(2022)Shi, Zhou, Hwang, Titsias, and
  Mackey]{shi2022gradient}
Shi, J., Zhou, Y., Hwang, J., Titsias, M., and Mackey, L.
\newblock Gradient estimation with discrete stein operators.
\newblock In \emph{NeurIPS}, 2022.

\bibitem[Tokui \& Sato(2017)Tokui and Sato]{Tokui2017EvaluatingTV}
Tokui, S. and Sato, I.
\newblock Evaluating the variance of likelihood-ratio gradient estimators.
\newblock In \emph{ICML}, 2017.

\bibitem[Tucker et~al.(2017)Tucker, Mnih, Maddison, Lawson, and
  Sohl-Dickstein]{Tucker2017REBARLU}
Tucker, G., Mnih, A., Maddison, C.~J., Lawson, J., and Sohl-Dickstein, J.~N.
\newblock Rebar: Low-variance, unbiased gradient estimates for discrete latent
  variable models.
\newblock In \emph{NIPS}, 2017.

\bibitem[van~den Oord et~al.(2017)van~den Oord, Vinyals, and
  Kavukcuoglu]{Oord2017NeuralDR}
van~den Oord, A., Vinyals, O., and Kavukcuoglu, K.
\newblock Neural discrete representation learning.
\newblock In \emph{NIPS}, 2017.

\bibitem[Weaver \& Tao(2001)Weaver and Tao]{Weaver2001TheOR}
Weaver, L. and Tao, N.
\newblock The optimal reward baseline for gradient-based reinforcement
  learning.
\newblock In \emph{UAI}, 2001.

\bibitem[Williams(1992)]{Williams1992SimpleSG}
Williams, R.~J.
\newblock Simple statistical gradient-following algorithms for connectionist
  reinforcement learning.
\newblock \emph{Machine Learning}, 8:\penalty0 229--256, 1992.

\bibitem[Xiao et~al.(2017)Xiao, Rasul, and Vollgraf]{Xiao2017FashionMNISTAN}
Xiao, H., Rasul, K., and Vollgraf, R.
\newblock Fashion-mnist: a novel image dataset for benchmarking machine
  learning algorithms.
\newblock \emph{ArXiv}, abs/1708.07747, 2017.

\end{thebibliography}
\bibliographystyle{icml2022}

\newpage
\appendix
\section{Theorem~\ref{theorem: st}}
\label{appendix:proof-st}
Let us define the first-order approximation of $\nabla$ as $\appnabla_{\mbox{\small 1st-order}} =  \sum_i \sum_j  \vpi_j \frac{\partial f(\mI_j)}{\partial \mI_j} (\mI_i - \mI_j) \frac{d\,\vpi_i}{d\,\vtheta}$, which approximates $f(\mI_i) - f(\mI_j)$ in Equation~\ref{eqn:exact_baseline_final} as $\frac{\partial f(\mI_j)}{\partial \mI_j} (\mI_i - \mI_j)$. 
\begin{customthm}{3.1}
\begin{eqnarray}
    E[\appnabla_{\mbox{\scriptsize ST}}] = \appnabla_{\mbox{\small 1st-order}}. 
    \nonumber
\end{eqnarray}
\end{customthm}
\begin{proof}
Based on the definition, we have:
\begin{eqnarray}
    \appnabla_{\mbox{\small 1st-order}} &=& \sum_i \sum_j  \vpi_j \frac{\partial f(\mI_j)}{\partial \mI_j} (\mI_i - \mI_j) \frac{d\,\vpi_i}{d\,\vtheta} \nonumber\\
    &=& \sum_j  \vpi_j \frac{\partial f(\mI_j)}{\partial \mI_j} \sum_i \mI_i \frac{d\,\vpi_i}{d\,\vtheta} - \sum_j  \vpi_j \frac{\partial f(\mI_j)}{\partial \mI_j} \mI_j \sum_i\frac{d\,\vpi_i}{d\,\vtheta} \label{eqn:theorey-st-eq1}
\end{eqnarray}
Since $\sum_i \vpi_i = 1$, we have $\sum_i\frac{d\,\vpi_i}{d\,\vtheta} = 0$. 
Also, since $\vpi = \sum_i \vpi_i \mI_i$, we have $\frac{d\,\vpi}{d\,\vtheta} = \sum_i \mI_i \frac{d\,\vpi_i}{d\,\vtheta}$. 
Thus, together with Equation~\ref{eqn:theorey-st-eq1}, we have:
\begin{eqnarray}
    \appnabla_{\mbox{\small 1st-order}} &=& \sum_j  \vpi_j \frac{\partial f(\mI_j)}{\partial \mI_j} \sum_i \mI_i \frac{d\,\vpi_i}{d\,\vtheta}\nonumber\\
    &=& E[\frac{\partial f(\mD)}{\partial \mD} \frac{d\,\vpi}{d\,\vtheta}] = E[\appnabla_{\mbox{\scriptsize ST}}]. \nonumber
\end{eqnarray}
\end{proof}

\section{Theorem~\ref{theorem: reinmax}}
\label{appendix:proof-reinmax}

\begin{customthm}{3.2}
\begin{eqnarray}
    E[\appnabla_{\mbox{\scriptsize ReinMax}}] = \appnabla_{\mbox{\small 2nd-order}}.
    \nonumber
\end{eqnarray}
\end{customthm}
\begin{proof}
Here, we aim to proof, $\forall k \in [1, n]$, we have $E[\appnabla_{\mbox{\scriptsize ReinMax}, k}] = \appnabla_{\mbox{\scriptsize 2nd-order}, k}$.
As defined in Equation~\ref{eqn: exact-nobaseline}, we have (note that $\delta_{\mD \mI_k}$ is the indicator function of the event $\mD = \mI_k$):
\begin{eqnarray}
\appnabla_{\mbox{\small 2nd-order}, k} &=& \sum_i \sum_j \frac{\vpi_j }{2} (\frac{\partial f(\mI_j)}{\partial \mI_j} + \frac{\partial f(\mI_i)}{\partial \mI_i} ) (\mI_i - \mI_j) \frac{d\,\vpi_i}{d\,\vtheta_k} \nonumber\\
&=& \sum_i \sum_j \frac{\vpi_j \vpi_i (\delta_{ik} - \vpi_k) }{2} (\frac{\partial f(\mI_j)}{\partial \mI_j} + \frac{\partial f(\mI_i)}{\partial \mI_i} ) (\mI_i - \mI_j) \nonumber\\
&=& \sum_j \frac{\vpi_j \vpi_k }{2} (\frac{\partial f(\mI_j)}{\partial \mI_j} + \frac{\partial f(\mI_k)}{\partial \mI_k} ) (\mI_k - \mI_j) \nonumber\\
&=& \frac{\vpi_k }{2} \frac{\partial f(\mI_k)}{\partial \mI_k} (\mI_k  -\sum_j  \vpi_j \mI_j) + \sum_j \frac{\vpi_j \vpi_k }{2} \frac{\partial f(\mI_j)}{\partial \mI_j} (\mI_k - \mI_j) \nonumber\\
&=& \frac{1}{2} E_{\mD\sim\vpi}[\delta_{\mD \mI_k} \frac{\partial f(\mD)}{\partial \mD} (\mI_{\mD} - \sum_j  \vpi_j \mI_j)] + \frac{1}{2} E_{\mD\sim\vpi}[\vpi_k \frac{\partial f(\mD)}{\partial \mD} (\mI_k - \mI_{\mD})] \nonumber\\
&=& \frac{1}{2} E_{\mD\sim\vpi}\large[\normalsize 
    \frac{\partial f(\mD)}{\partial \mD} 
        \large(\normalsize
        \vpi_k (\mI_k - \mI_{\mD}) + \delta_{\mD \mI_k} (\mI_{\mD} - \sum_i  \vpi_i \mI_i)
        \large)\normalsize
    \large]\normalsize 
    \label{eqn:theorey-reinmax-right}
\end{eqnarray}
At the same time, based on the definition of $\appnabla_{\mbox{\scriptsize ReinMax}}$, we have:
\begin{eqnarray}
E[\appnabla_{\mbox{\scriptsize ReinMax}, k}] &=& E_{\mD\sim\vpi}\large[\normalsize
    \frac{\partial f(\mD)}{\partial \mD} 
        \large(\normalsize
        2 \cdot \frac{\vpi_k + \delta_{\mD \mI_k}}{2} (\mD_k - \sum_i \frac{\vpi_i + \delta_{\mD \mI_k}}{2} \mI_i) - \frac{ \vpi_k}{2} (\mD_k - \sum_i \vpi_i \mI_i)
        \large)\normalsize
    \large]\normalsize  \nonumber\\
    &=& \frac{1}{2} E_{\mD\sim\vpi}\large[\normalsize 
    \frac{\partial f(\mD)}{\partial \mD} 
        \large(\normalsize
        \vpi_k (\mI_k - \mI_{\mD}) + \delta_{\mD \mI_k} (\mI_{k} - \sum_i  \vpi_i \mI_i)
        \large)\normalsize
    \large]\normalsize 
    \label{eqn:theorey-reinmax-left}
\end{eqnarray}
Since $\delta_{\mD \mI_k} (\mI_{k} - \sum_i  \vpi_i \mI_i) = \delta_{\mD \mI_k} (\mI_{\mD} - \sum_i  \vpi_i \mI_i)$, together with Equation~\ref{eqn:theorey-reinmax-right} and \ref{eqn:theorey-reinmax-left}, we have:
\begin{eqnarray}
E[\appnabla_{\mbox{\scriptsize ReinMax}, k}] = \appnabla_{\mbox{\small 2nd-order}, k}
\nonumber
\end{eqnarray}
\end{proof}

\section{Remark~\ref{theorem: st-avg}}
\label{appendix:proof-st-avg}

\begin{customrmk}{3.1}
When $\sum_i \vphi_i f(\mI_i)$ is used as the baseline and  $f(\mI_i) - f(\mI_j)$ is approximated as $\frac{\partial f(\mI_j)}{\partial \mI_j} (\mI_i - \mI_j)$, we mark the resulting first-order approximation of $\nabla$ as $\appnabla_{\mbox{\small 1st-order-avg-baseline}}$.
Then, we have:
\begin{eqnarray}
    E[\frac{\vphi_\mD}{\vpi_\mD}\appnabla_{\mbox{\scriptsize ST}}] = \appnabla_{\mbox{\small 1st-order-avg-baseline}}
    \nonumber
\end{eqnarray}
\end{customrmk}
\begin{proof}
Using $\sum_i \vphi_i f(\mI_i)$ as the baseline, we have:
\begin{eqnarray*}
\nabla =  \sum_i (f(\mI_i) - \sum_j \vphi_j f(\mI_j)) \frac{d\,\vpi_i}{d\,\vtheta} = \sum_i \sum_j  \vphi_j (f(\mI_i) - f(\mI_j)) \frac{d\,\vpi_i}{d\,\vtheta}
\end{eqnarray*}
Approximating $f(\mI_i) - f(\mI_j)$ as $\frac{\partial f(\mI_j)}{\partial \mI_j} (\mI_i - \mI_j)$, we have:
\begin{eqnarray*}
\appnabla_{\mbox{\small 1st-order-avg-baseline}} &=& \sum_i \sum_j  \vphi_j \frac{\partial f(\mI_j)}{\partial \mI_j} (\mI_i - \mI_j) \frac{d\,\vpi_i}{d\,\vtheta} \\
&=& \sum_j  \frac{\vphi_j}{\vpi_j} \cdot \vpi_j \cdot \frac{\partial f(\mI_j)}{\partial \mI_j}\sum_i  \mI_i  \frac{d\,\vpi_i}{d\,\vtheta} \\
&=& E[\frac{\vphi_\mD}{\vpi_\mD}\appnabla_{\mbox{\scriptsize ST}}] 
\end{eqnarray*}
\end{proof}

\section{Remark~\ref{theorem: reinmax-nobaseline}}
\label{appendix:proof-reinmax-nobaseline}

\begin{customrmk}{3.2}
In Equation~\ref{eqn: exact-nobaseline}, we approximate $f(\mI_k) - f(\mI_i)$ as $\frac{1}{2} (\frac{\partial f(\mI_i)}{\partial \mI_i} + \frac{\partial f(\mI_k)}{\partial \mI_k} )(\mI_k - \mI_i)$, and mark the resulting second-order approximation of $\nabla_k$ as $\appnabla_{\mbox{\small 2nd-order-wo-baseline}, k} =  \vpi_k \sum_i \vpi_i \frac{1}{2} (\frac{\partial f(\mI_i)}{\partial \mI_i} + \frac{\partial f(\mI_k)}{\partial \mI_k} ) (\mI_k - \mI_i)$,
Then, we have:
\begin{eqnarray}
    E[\appnabla_{\mbox{\scriptsize ReinMax}}] = \appnabla_{\mbox{\small 2nd-order-wo-baseline}}
    \nonumber
\end{eqnarray}
\end{customrmk}
\begin{proof}
Here, we aim to proof, $\forall k \in [1, n]$, we have $E[\appnabla_{\mbox{\scriptsize ReinMax}, k}] = \appnabla_{\mbox{\small 2nd-order-wo-baseline}, k}.$

\begin{equation*}
\begin{split}
\appnabla_{\mbox{\small 2nd-order-wo-baseline}, k} &=  \vpi_k \sum_i \vpi_i \frac{1}{2} (\frac{\partial f(\mI_i)}{\partial \mI_i} + \frac{\partial f(\mI_k)}{\partial \mI_k} ) (\mI_k - \mI_i) \\
 &= \vpi_k \sum_i \vpi_i \frac{1}{2} \frac{\partial f(\mI_i)}{\partial \mI_i} (\mI_k - \mI_i) + \vpi_k \sum_i \vpi_i \frac{1}{2} \frac{\partial f(\mI_k)}{\partial \mI_k}  (\mI_k - \mI_i) \\
 &=  E\large[\normalsize 
    \frac{\partial f(\mD)}{\partial \mD}
    \frac{\vpi_k (\mI_k - \mI_{\mD}) + \delta_{\mD \mI_k} (\mI_{k} - \sum_i  \vpi_i \mI_i)}{2}
    \large]\normalsize = E[\appnabla_{\mbox{\scriptsize ReinMax}, k}]
\end{split}
\end{equation*}
\end{proof}

\section{Forward Euler Method and Heun's Method}
\label{appendix:ode}

For simplicity, we consider a simple function $g(x): \mathcal{R} \to \mathcal{R}$ that is three times differentiable on $[t_0, t_1]$. 
Now, we proceed to a simple introduction to approximate $\int_{t_0}^{t_1} g'(x) dx$ with the Forward Euler Method and the Heun's Method. 
For a detailed introduction to numerical ODE methods, please refer to \citet{Ascher1998ComputerMF}. 

\smallsection{Forward Euler Method}
Here, we approximate $g(t_1)$ with the first-order Taylor expansion, i.e.,  $g(t_1) = g(t_0) + g'(t_0) \cdot (t_1- t_0) + O((t_1 - t_0)^2)$, then we have $\int_{t_0}^{t_1} g'(x) dx \approx g'(t_0) (t_1 - t_0)$. 
Since we used the first-order Taylor expansion, this approximation has first-order accuracy. 

\smallsection{Heun's Method}
First, we approximate $g(t_1)$ with the second-order Taylor expansion: 
\begin{equation}
g(t_1) = g(t_0) + g'(t_0) \cdot (t_1 - t_0) + \frac{g''(t_0)}{2} \cdot (t_1 - t_0)^2 + O((t_1 - t_0)^3).
\label{eqn:taylor-2nd}
\end{equation}
Then, we show that we can match this approximation by combining the first-order derivatives of two samples. 
Taylor expanding $g'(t_1)$ to the first-order, we have:
\begin{equation}
g'(t_1) = g'(t_0) + g''(t_0) \cdot (t_1 - t_0) + O((t_1 - t_0)^2) 
\nonumber
\end{equation}
Therefore, we have:
\begin{equation}
g(t_0) + \frac{g'(t_0) + g'(t_1)}{2} (t_1 - t_0) = g(t_0) + g'(t_0) \cdot (t_1 - t_0) + \frac{g''(t_0)}{2} \cdot (t_1 - t_0)^2 + O((t_1 - t_0)^3).
\nonumber
\end{equation}
It is easy to notice that the right-hand side of the above equation matches the second-order Taylor expansion of $g(t_1)$ as in Equation~\ref{eqn:taylor-2nd}.  
Therefore, the above approximation (i.e., approximating $g(t_1) - g(t_0)$ as $\frac{g'(t_0) + g'(t_1)}{2} (t_1 - t_0)$) has second-order accuracy.

\smallsection{Connection to $f(\mI_i) - f(\mI_j)$ in Equation~\ref{eqn:exact_baseline_final}}
By setting $g(x) = f(x \cdot \mI_i + (1 -x) \cdot \mI_j))$, we have $g(1) - g(0) = f(\mI_i) - f(\mI_j)$. 
Then, it is easy to notice that the forward Euler Method approximates $f(\mI_i) - f(\mI_j)$ as $\frac{\partial f(\mI_j)}{\partial \mI_j} (\mI_i - \mI_j)$ and has first-order accuracy. 
Also, the Heun's Method approximates $f(\mI_i) - f(\mI_j)$ as $\frac{1}{2} (\frac{\partial f(\mI_i)}{\partial \mI_i} + \frac{\partial f(\mI_j)}{\partial \mI_j})(\mI_i - \mI_j)$ and has second-order accuracy. 


\section{Experiment Details}
\label{appendix:exp}

\subsection{Baselines}
Here, we consider four methods as our major baselines:
\begin{itemize}[leftmargin=*]
    \item
    \vspace{-0.1in}
    Straight-Through (ST; \citeauthor{Bengio2013EstimatingOP}, \citeyear{Bengio2013EstimatingOP}) backpropagate through the sampling function as if it had been the identity function. 

    \item
    Straight-Through Gumbel-Softmax (STGS; \citeauthor{Jang2016CategoricalRW}, \citeyear{Jang2016CategoricalRW}) integrates the Gumbel reparameterization trick to approximate the gradient. 

    \item
    Gumbel-Rao Monte Carlo (GR-MCK; \citeauthor{Paulus2020RaoBlackwellizingTS}, \citeyear{Paulus2020RaoBlackwellizingTS}) leverages the Monte Carlo method to reduce the variance introduced by the Gumbel noise in STGS. To obtain the optimal performance for this baseline, we set the number of Monte Carlo samples to 1000 in most experiments. Except in our discussions of efficiency, we set the number of Monte Carlo samples to 100, 300, and 1000 for a more comprehensive comparisons. 

    \item
    Gapped Straight-Through (GST-1.0; \citeauthor{Fan2022TrainingDD}, \citeyear{Fan2022TrainingDD}) aims to reduce the variance of STGS and constructs a deterministic term to replace the Monte Carlo samples used in GR-MCK.
    Here, as suggested in \citep{Fan2022TrainingDD}, we set the gap (a hyper-parameter) as $1.0$.
\end{itemize}

\smallsection{GST-1.0 Performance}
Despite GST-1.0 achieving good performance on most settings of MNIST-VAE, it fails to maintain this performance on polynomial programming and unsupervised parsing, as discussed before. 
At the same time, 
a different variant of GST (i.e., GST-p) achieves a significant performance boost over GST-1.0 on polynomial programming. 
However, on MNIST-VAE and ListOps, GST-p achieves an inferior performance. 
Upon discussing with the author of the GST-1.0, we suggest that this phenomenon is caused by different characteristics of GST-1.0 and GST-p. 

This observation verifies our intuition that, without understanding the mechanism of ST, different applications have different preferences on its configurations.
Meanwhile, ReinMax achieves consistent improvements in all settings, which greatly simplifies future algorithms developments.

\subsection{Hyper-Parameters}
Without specifically, we conduct full grid search for all methods in all experiments, and report the best performance (averaged with 10 random seeds on MNIST-VAE and 5 random seeds on ListOps). 
The hyper-parameter search space is summarized in Table~\ref{table:hyperparameter}. 
The search results for Table~\ref{table:mnist-vae} and Table~\ref{table:listops} are summarized in Table~\ref{table:opt-hyperparameter}.

\begin{table}[htbp]
\caption{Hyper-parameter search space.}
\centering
\begin{tabular}{lc}
\toprule
\textbf{Hyperparameters} & Search Space\\
\midrule
Optimizer & \{Adam\citep{Kingma2014AdamAM}, RAdam\citep{Liu2019OnTV}\} \\
Learning Rate & \{0.001, 0.0007, 0.0005, 0.0003\} \\
Temperature & \{0.1, 0.3, 0.5, 0.7, 1.0, 1.1, 1.2, 1.3, 1.4, 1.5\}  \\
\bottomrule
\end{tabular}
\label{table:hyperparameter}
\end{table}
\begin{table}[htbp]
\caption{Hyper-parameters Search Result for Results in Table~\ref{table:listops} and Table~\ref{table:mnist-vae}.}
\centering
\scalebox{0.95}{
\begin{tabular}{ll|ccccc}
\toprule
 &  & STGS & GR-MCK & GST-1.0 & ST & \ours \\
\midrule
                          & Optimizer & Adam & Adam & Adam & Adam & Adam
                          \\\cmidrule{2-7}
MNIST-VAE $8 \times 4$    & Learning Rate & 0.0003 & 0.0005 & 0.0005 & 0.001 & 0.0005
                          \\\cmidrule{2-7}
                          & Temperature & 0.5 & 0.5 & 0.7 & 1.3 & 1.3
                          \\\midrule
                          & Optimizer & RAdam & RAdam & RAdam & RAdam & RAdam
                          \\\cmidrule{2-7}
MNIST-VAE $4 \times 24$   & Learning Rate & 0.0005 & 0.0005 & 0.0005 & 0.001 & 0.0005
                          \\\cmidrule{2-7}
                          & Temperature & 0.3 & 0.3 & 0.5 & 1.5 & 1.5
                          \\\midrule
                          & Optimizer & RAdam & RAdam & RAdam & RAdam & RAdam
                          \\\cmidrule{2-7}
MNIST-VAE $8 \times 16$   & Learning Rate & 0.0005 & 0.0007 & 0.0007 & 0.001 & 0.0007
                          \\\cmidrule{2-7}
                          & Temperature & 0.5 & 0.7 & 0.5 & 1.5  & 1.5
                          \\\midrule
                          & Optimizer & RAdam & Adam & RAdam & Adam & RAdam
                          \\\cmidrule{2-7}
MNIST-VAE $16 \times 12$  & Learning Rate & 0.0007 & 0.0005 & 0.0007 & 0.0005 & 0.0007
                          \\\cmidrule{2-7}
                          & Temperature & 0.7 & 1.0 & 0.5 & 1.5 & 1.5
                          \\\midrule
                          & Optimizer & RAdam & Adam & RAdam & Adam & RAdam
                          \\\cmidrule{2-7}
MNIST-VAE $64 \times 8$   & Learning Rate & 0.0007 & 0.0007 & 0.0007 & 0.0005 & 0.0005
                          \\\cmidrule{2-7}
                          & Temperature & 0.7 & 2.0 & 0.7 & 1.5 & 1.5
                          \\\midrule
                          & Optimizer & RAdam & RAdam& RAdam & RAdam & RAdam
                          \\\cmidrule{2-7}
MNIST-VAE $10 \times 30$  & Learning Rate & 0.0005 & 0.0005 & 0.0005 & 0.0007 & 0.0005
                          \\\cmidrule{2-7}
                          & Temperature & 0.5 & 1.0 & 0.5 & 1.4 & 1.3
                          \\\midrule
                          & Optimizer & RAdam & RAdam & RAdam & RAdam & RAdam
                          \\\cmidrule{2-7}
ListOps                   & Learning Rate & 0.0005 & 0.0005 & 0.001 & 0.001 & 0.0007
                          \\\cmidrule{2-7}
                          & Temperature & 0.1 & 0.3 & 0.1 & 1.4 & 1.1
                          \\
\bottomrule
\end{tabular}
}
\label{table:opt-hyperparameter}
\end{table}


\smallsection{Polynomial Programming}
As this problem is relatively simple, we set the learning rate to 0.001 and the optimizer to Adam, and only tune the temperature hyper-parameter. 

\smallsection{MNIST-VAE} Following the previous study~\citep{Dong2020DisARMAA,Dong2021CoupledGE,Fan2022TrainingDD}, we used 2-layer MLP as the encoder and the decoder. We set the hidden state dimension of the first-layer and the second-layer as $512$ and $256$ for the encoder, and $256$ and $512$ for the decoder. 
For our experiments on MNIST-VAE with 32 latent dimensions and 64 categorical dimensions, we set the batch size to 200, training steps to $5\times 10^5$, and activation function to LeakyReLU, in order to be consistent with the literature. 
For other experiments, we set the batch size to 100, the activation function to ReLU, and training steps to $9.6 \times 10^4$ (i.e., 160 epochs). 

\smallsection{Differentiable Neural Architecture Search}
We adopt most of the hyper-parameter setting from \citet{Dong2020NATSBenchBN}. 
Since GDAS employs a temperature schedule (decaying linearly from 10 to 0.1), and temperature scaling works differently in \ours and STGS (as discussed in Section~\ref{sec:temperature} and Section~\ref{subsec: exp-discussions}), we removed the temperature scaling (i.e., set the temperature to a constant $1.0$) and increased the weight decay (i.e., from $0.001$ to $0.09$).

\begin{table}[t!]

\centering
\caption{Test $-$ELBO on MNIST. Hyper-parameters are chosen based on Train $-$ELBO.}
\label{table:mnist-vae-test-overfitting}
\scalebox{.83}{
\begin{tabular}{l|c|c|cccc|c}
\toprule
  & AVG & $8\times 4$ & $4\times 24$ & $8\times 16$ & $16\times 12$ & $64\times 8$ & $10\times 30$ \\ 
\midrule
STGS    & 106.89 & 128.09$\pm$0.79 & 103.60$\pm$0.45 &  99.32$\pm$0.33 & 102.49$\pm$0.32 & 106.20$\pm$0.46 &  101.61$\pm$0.54 \\ 
GR-MCK  & 109.03 & 127.90$\pm$0.71 & 102.76$\pm$0.33 &  102.12$\pm$0.29 & 104.23$\pm$0.65 & 113.54$\pm$0.50 & 103.62$\pm$0.13 \\ 
GST-1.0 & 106.85 & 128.20$\pm$1.12 & 103.95$\pm$0.49 &  101.44$\pm$0.32 & 101.28$\pm$0.59 & 105.44$\pm$0.62 & 100.78$\pm$0.44 \\ 
\midrule
ST      & 118.85 & 137.06$\pm$0.51 & 113.41$\pm$0.49 & 114.25$\pm$0.29 & 114.48$\pm$0.56 & 115.43$\pm$0.29 & 118.46$\pm$0.18 \\
\ours   & \textbf{105.74} & \textbf{126.89$\pm$0.79} &  \textbf{102.40$\pm$0.43} &  \textbf{100.63$\pm$0.41} & \textbf{100.85$\pm$0.50} & \textbf{102.91$\pm$0.67} &  \textbf{100.75$\pm$0.50} \\
\bottomrule
\end{tabular}
}
\end{table}

\begin{table}[t]

\centering
\caption{Test $-$ELBO on MNIST. Hyper-parameters are chosen based on Test $-$ELBO.}
\label{table:mnist-vae-test-best}
\scalebox{.83}{
\begin{tabular}{l|c|c|cccc|c}
\toprule
  & AVG & $8\times 4$ & $4\times 24$ & $8\times 16$ & $16\times 12$ & $64\times 8$ & $10\times 30$ \\ 
\midrule
STGS    & 107.15 & 128.09$\pm$0.79 & 103.25$\pm$0.22 & 101.44$\pm$0.32 & 102.29$\pm$0.39 & 106.20$\pm$0.46 & 101.61$\pm$0.54 \\ 
GR-MCK  & 108.87 & 127.86$\pm$0.54 & 102.40$\pm$0.37 & 101.59$\pm$0.22 & 104.22$\pm$0.63 & 113.54$\pm$0.50 & 103.62$\pm$0.13 \\ 
GST-1.0 & 106.55 & 128.03$\pm$1.02 & 103.63$\pm$0.24 & 100.67$\pm$0.34 & 101.04$\pm$0.39 & 105.44$\pm$0.62 & \textbf{100.51$\pm$0.37} \\ 
\midrule
ST      & 118.79 & 137.05$\pm$0.36 & 113.23$\pm$0.43 & 114.11$\pm$0.31 & 114.48$\pm$0.56 & 115.43$\pm$0.29 & 118.46$\pm$0.18 \\
\ours   & \textbf{105.60} & \textbf{126.29$\pm$0.32} & \textbf{102.40$\pm$0.43} & \textbf{100.45$\pm$0.26} & \textbf{100.84$\pm$0.56} & \textbf{102.91$\pm$0.68} & 100.69$\pm$0.48 \\
\bottomrule
\end{tabular}
}
\end{table}

\begin{table}[t!]

\centering
\caption{Train $-$ELBO on MNIST. Hyper-parameters are chosen based on Test $-$ELBO.}
\label{table:mnist-vae-train-overfitting}
\scalebox{.83}{
\begin{tabular}{l|c|c|cccc|c}
\toprule
  & AVG & $8\times 4$ & $4\times 24$ & $8\times 16$ & $16\times 12$ & $64\times 8$ & $10\times 30$ \\ 
\midrule
STGS    & 105.31 & 126.85$\pm$0.85 & 101.81$\pm$0.14 &  99.32$\pm$0.33 & 100.22$\pm$0.47 & 104.02$\pm$0.41 &  99.63$\pm$0.63 \\ 
GR-MCK  & 107.37 & 126.53$\pm$0.55 & 100.47$\pm$0.31 &  99.75$\pm$0.29 & 103.11$\pm$0.58 & 112.34$\pm$0.48 & 102.02$\pm$0.18 \\ 
GST-1.0 & 104.60 & 126.63$\pm$1.16 & 102.11$\pm$0.24 &  98.40$\pm$0.34 & 98.76$\pm$0.41 & 102.53$\pm$0.57 &  99.14$\pm$0.30 \\ 
\midrule
ST      & 117.76 & 136.75$\pm$0.22 & 112.09$\pm$0.50 & 113.06$\pm$0.26 & 113.31$\pm$0.43 & 113.90$\pm$0.28 & 117.46$\pm$0.09 \\
\ours   & \textbf{103.40} & \textbf{124.92$\pm$0.38} &  \textbf{99.77$\pm$0.45} &  \textbf{98.06$\pm$0.31} & \textbf{98.51$\pm$0.54} & \textbf{100.71$\pm$0.70} &  \textbf{98.40$\pm$0.48} \\
\bottomrule
\end{tabular}
}
\end{table}

\smallsection{ListOps}
We followed the same setting of \citet{Fan2022TrainingDD}, i.e., used the same model configuration as in \citet{Choi2017LearningTC} and set the maximum sequence length to $100$.

\subsection{Hardware and Environment Setting}
Most experiments (except efficiency comparisons) are conducted on Nvidia P40 GPUs. 
For efficiency comparisons, 
we measured the average time cost per batch and peak memory consumption on quadratic programming and MNIST-VAE on the same system with an idle A6000 GPU. 
Also, to better reflect the efficiency of gradient estimators, we skipped all parameter updates in efficiency comparisons. 

\subsection{Additional Results on Polynomial Programming}
\begin{figure}[b!]
    \centering
    \includegraphics[width=1.\textwidth]{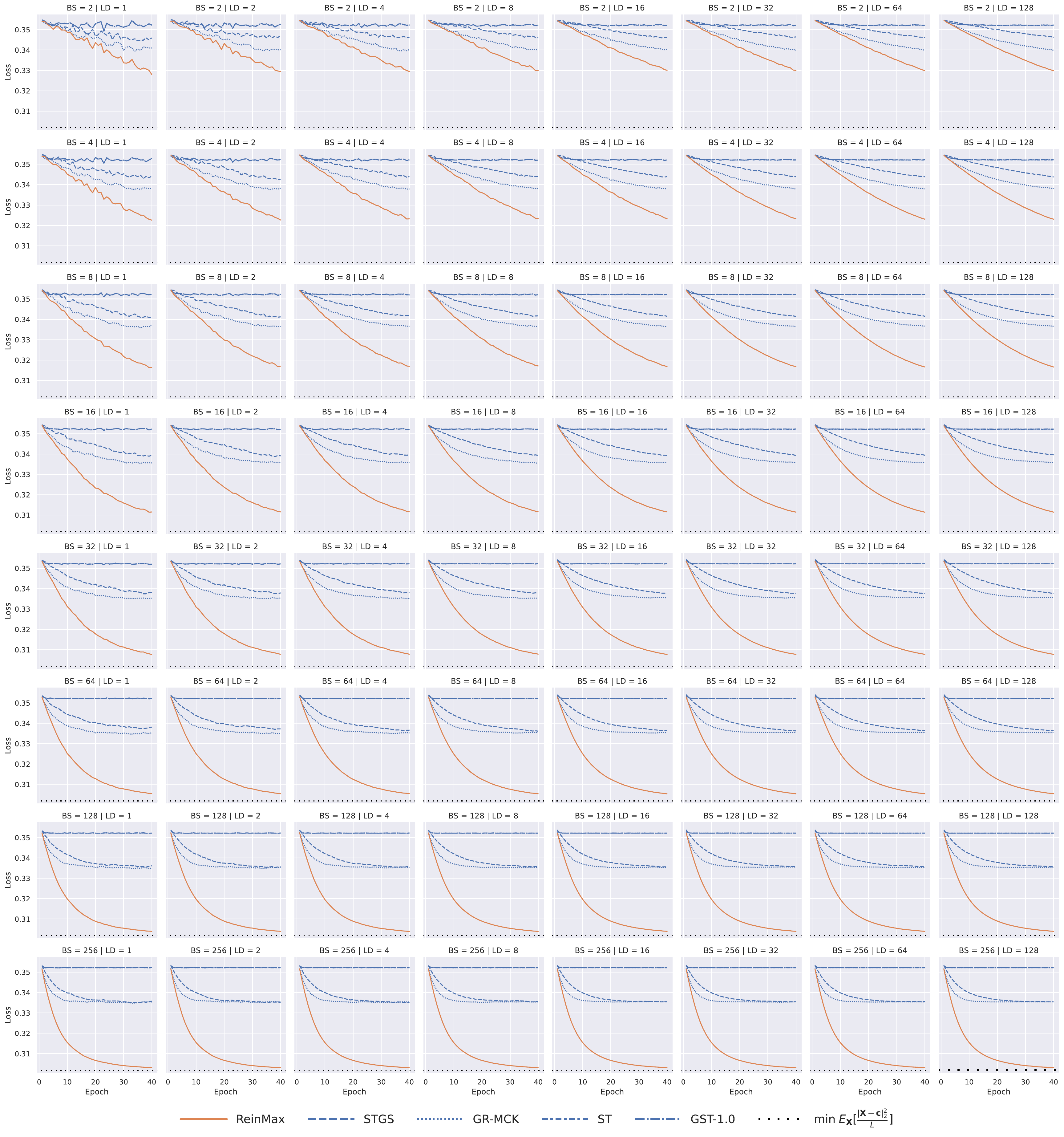}
    \vspace{-6mm}
    \caption{Polynomial programming training curve, with different batch sizes and random variable counts ($L$), i.e., $\min_\vtheta E\rrrl\frac{\|\mX - \vc\|_{1.5}^{1.5}}{L}\rrrr, \;\mbox{ where } \vtheta \in \mathcal{R}^{L\times 2}, \mX \in \{0, 1\}^{L}, \mbox{ and } \mX_i \overset{\mathrm{iid}}{\sim} \mbox{Multinomial}(\mbox{softmax}(\vtheta_i))$. More details are elaborated in Section~\ref{sect:experiment}.}
    \label{fig:poly_all_1.5}
\end{figure}

\begin{figure}[b!]
    \centering
    \includegraphics[width=1.\textwidth]{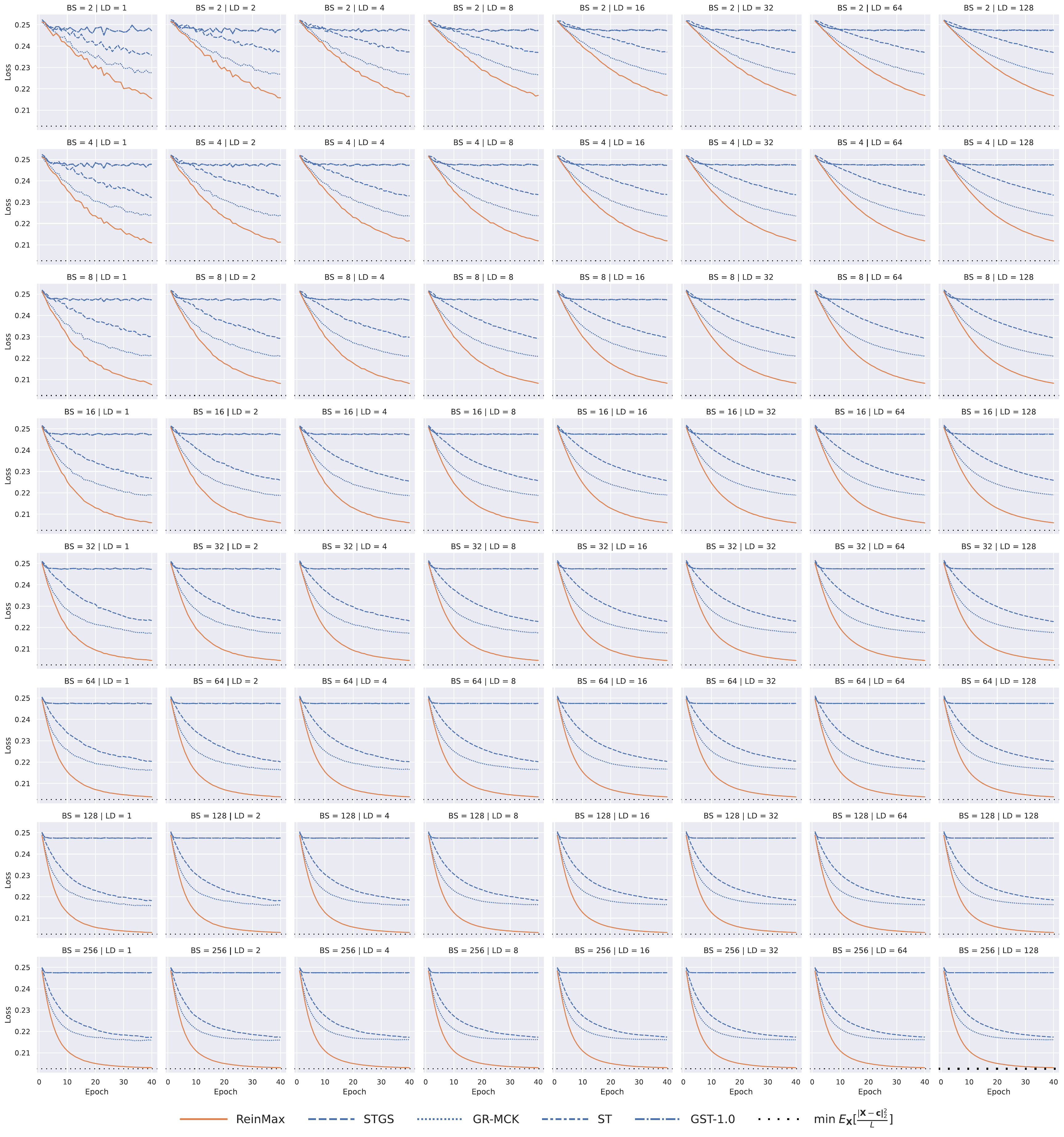}
    \vspace{-6mm}
    \caption{Quadratic programming training curve, with different batch sizes and random variable counts ($L$), i.e., $\min_\vtheta E \rrrl \frac{\|\mX - \vc\|_2^2}{L}\rrrr, \;\mbox{ where } \vtheta \in \mathcal{R}^{L\times 2}, \mX \in \{0, 1\}^{L}, \mbox{ and } \mX_i \overset{\mathrm{iid}}{\sim} \mbox{Multinomial}(\mbox{softmax}(\vtheta_i))$. More details are elaborated in Section~\ref{sect:experiment}.}
    \label{fig:poly_all_2}
\end{figure}

\begin{figure}[b!]
    \centering
    \includegraphics[width=1.\textwidth]{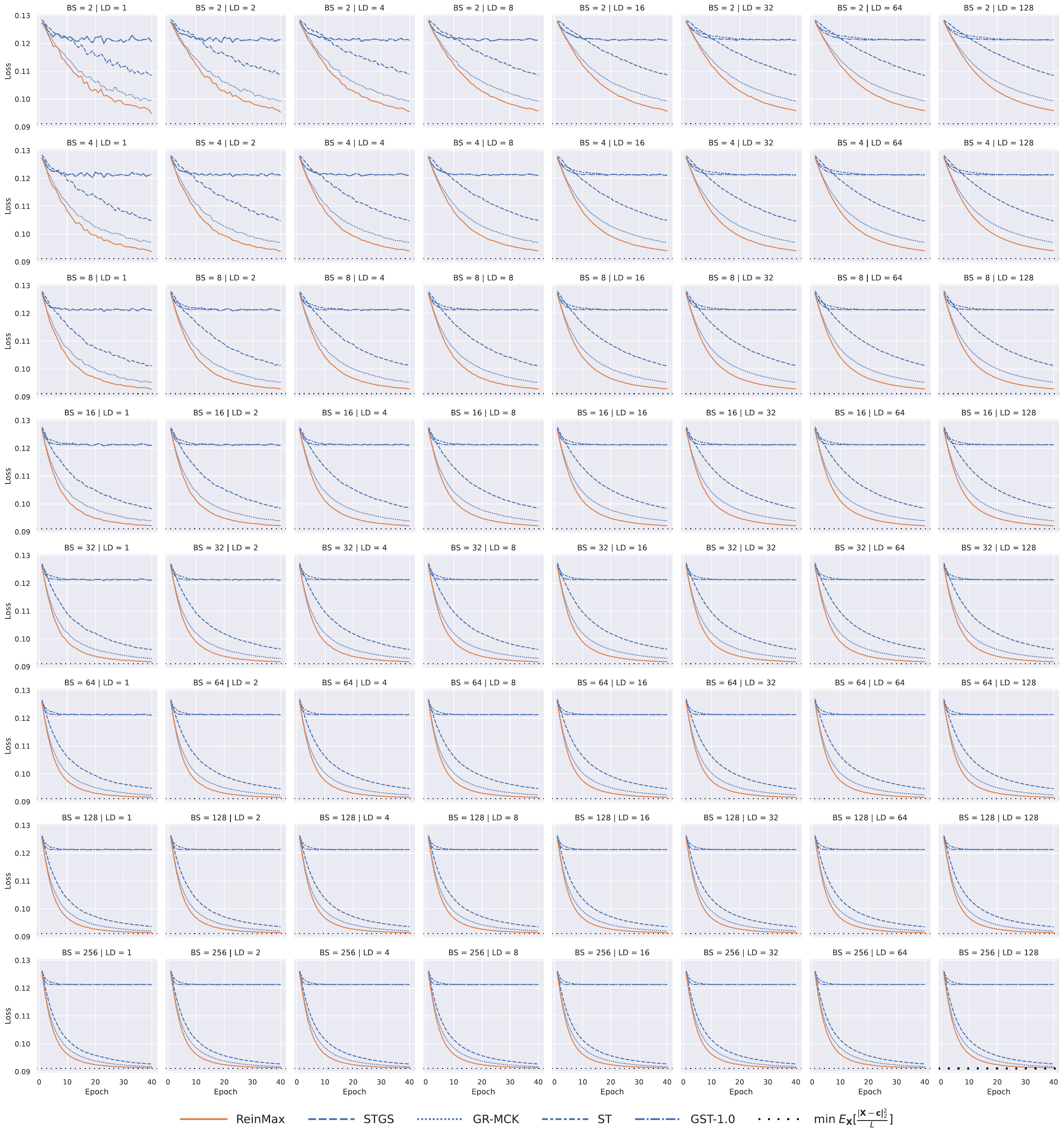}
    \vspace{-6mm}
    \caption{Polynomial programming training curve, with different batch sizes and random variable counts ($L$), i.e., $\min_\vtheta E \rrrl \frac{\|\mX - \vc\|_3^3}{L}\rrrr, \;\mbox{ where } \vtheta \in \mathcal{R}^{L\times 2}, \mX \in \{0, 1\}^{L}, \mbox{ and } \mX_i \overset{\mathrm{iid}}{\sim} \mbox{Multinomial}(\mbox{softmax}(\vtheta_i))$. More details are elaborated in Section~\ref{sect:experiment}.}
    \label{fig:poly_all_3}
\end{figure}

We visualized the training curve for polynomial programming with various batch sizes and latent dimensions in Figure~\ref{fig:poly_all_1.5} (for $p=1.5$), Figure~\ref{fig:poly_all_2} (for $p=2$), and Figure~\ref{fig:poly_all_3} (for $p=3$).

\begin{figure}[h!]
    \centering
    \begin{subfigure}[t]{0.48\linewidth}
        \centering
        \includegraphics[width=1.0\textwidth]{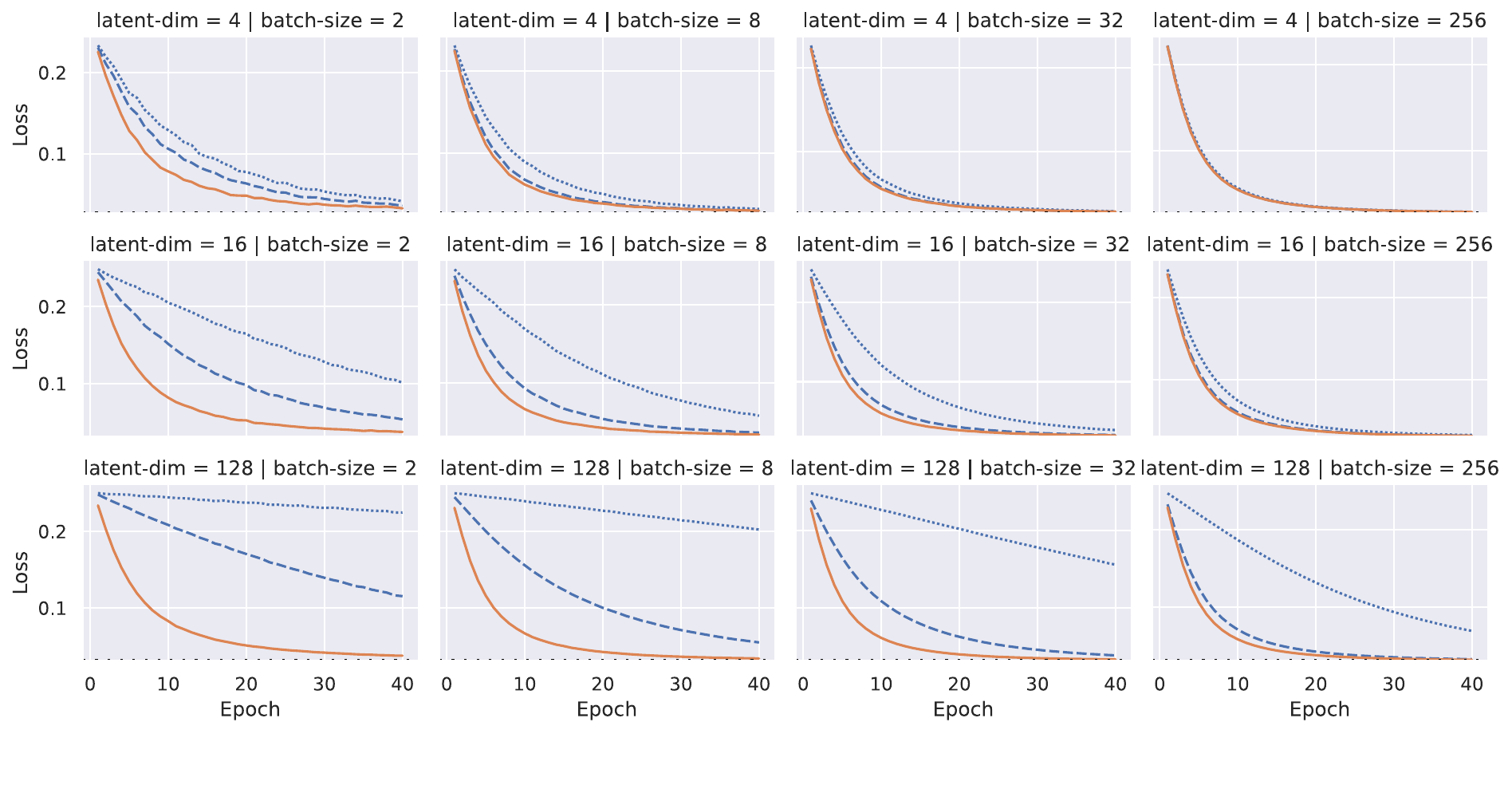}
        \vspace{-8mm}
        \caption{$p=3$ and $\vc = \rrrl\frac{0.5}{L}, \frac{1.5}{L}, \cdots, \frac{L-0.5}{L}\rrrr$.}
        \label{}
    \end{subfigure}
    \hfill
    \begin{subfigure}[t]{0.48\linewidth}
        \centering
        \includegraphics[width=1.0\textwidth]{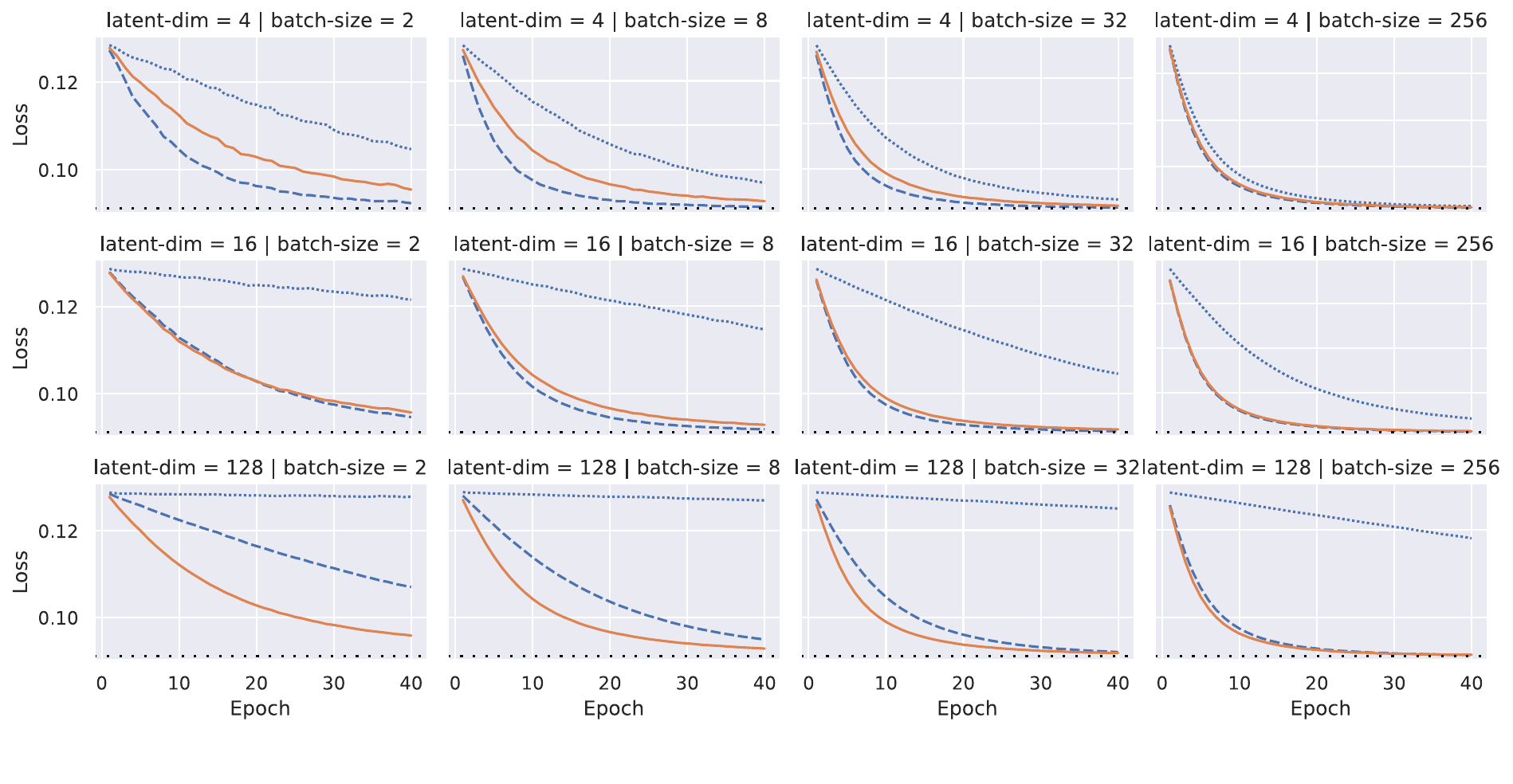}
        \vspace{-8mm}
        \caption{$p=3$ and $\vc = \rrrl0.45, \cdots, 0.45\rrrr$.}
        \label{}
    \end{subfigure}
    
    \vfill
    
    \begin{subfigure}[t]{0.48\linewidth}
        \centering
        \includegraphics[width=1.0\textwidth]{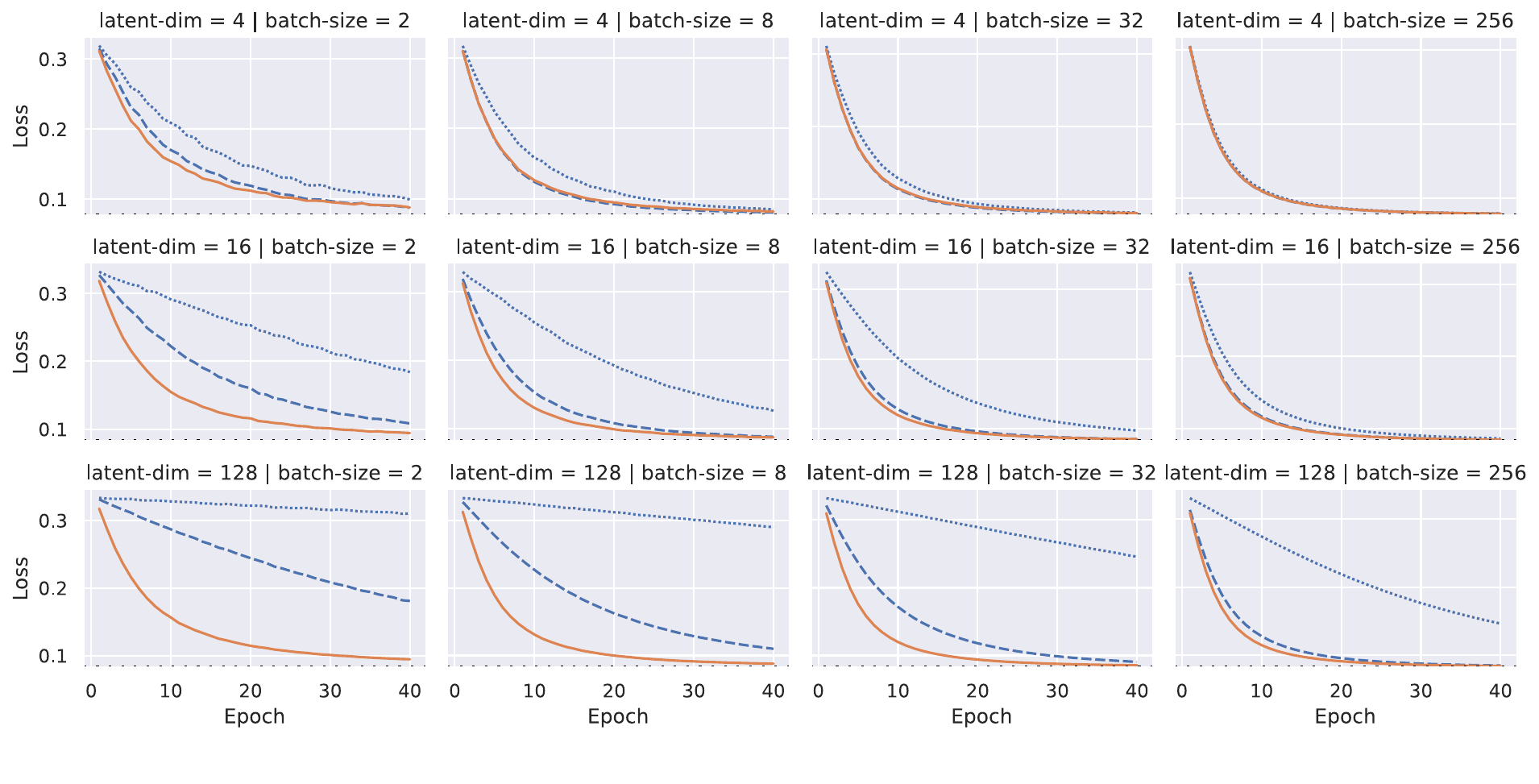}
        \vspace{-8mm}
        \caption{$p=2$ and $\vc = \rrrl\frac{0.5}{L}, \frac{1.5}{L}, \cdots, \frac{L-0.5}{L}\rrrr$.}
        \label{}
    \end{subfigure}
    \hfill
    \begin{subfigure}[t]{0.48\linewidth}
        \centering
        \includegraphics[width=1.0\textwidth]{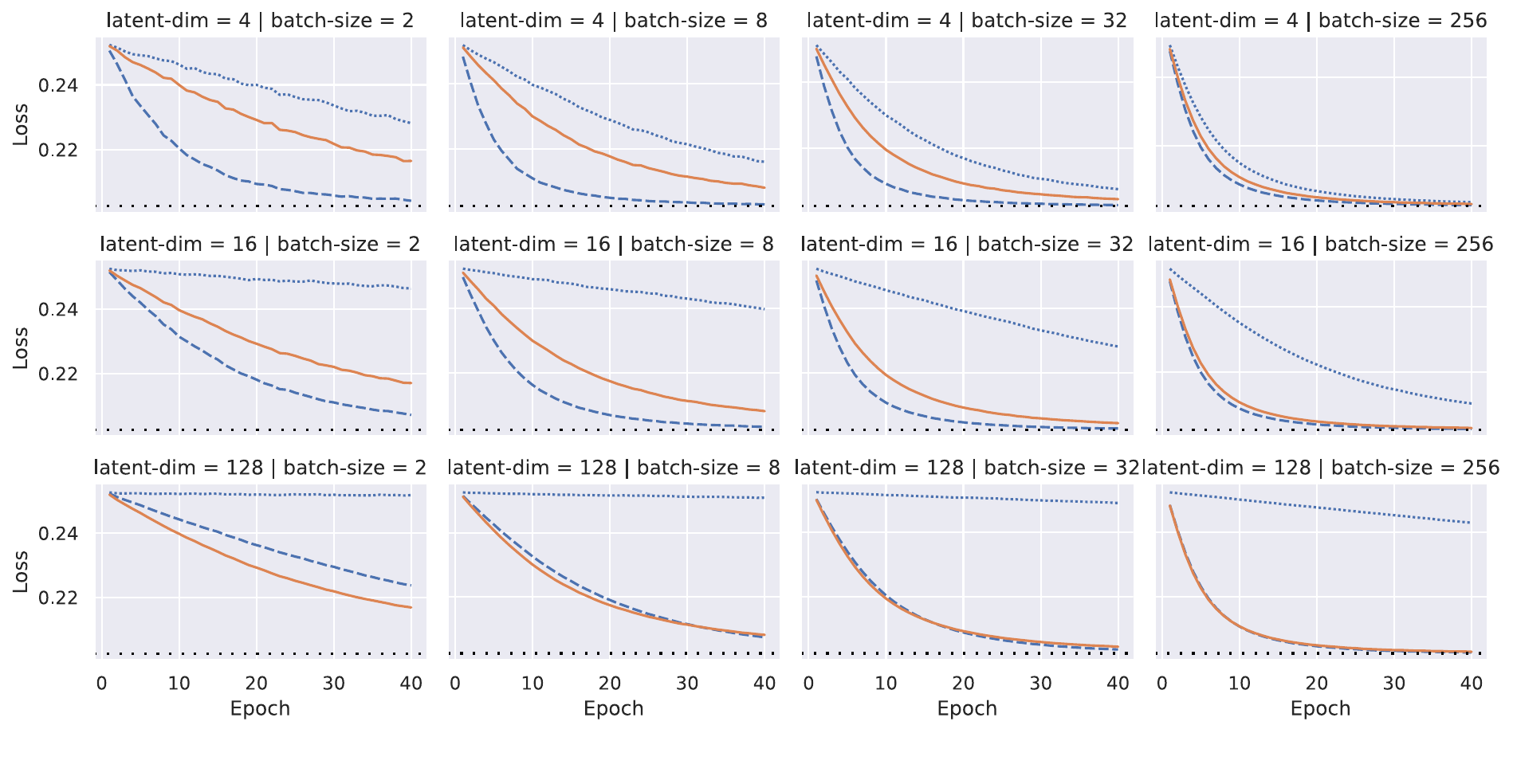}
        \vspace{-8mm}
        \caption{$p=2$ and $\vc = \rrrl0.45, \cdots, 0.45\rrrr$.}
        \label{}
    \end{subfigure}

    \vfill
    
    \begin{subfigure}[t]{0.48\linewidth}
        \centering
        \includegraphics[width=1.0\textwidth]{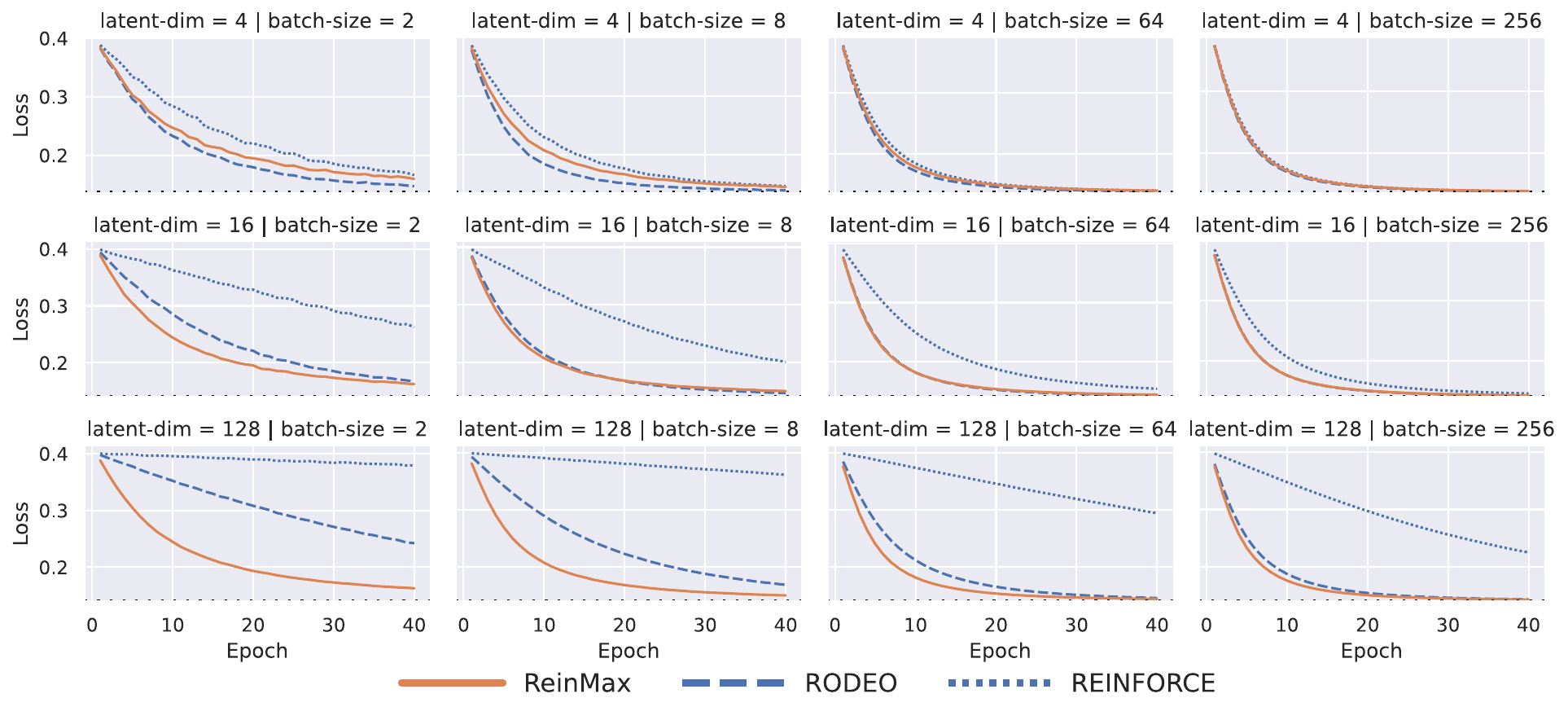}
        \vspace{-4mm}
        \caption{$p=1.5$ and $\vc = \rrrl\frac{0.5}{L}, \frac{1.5}{L}, \cdots, \frac{L-0.5}{L}\rrrr$.}
        \label{}
    \end{subfigure}
    \hfill
    \begin{subfigure}[t]{0.48\linewidth}
        \centering
        \includegraphics[width=1.0\textwidth]{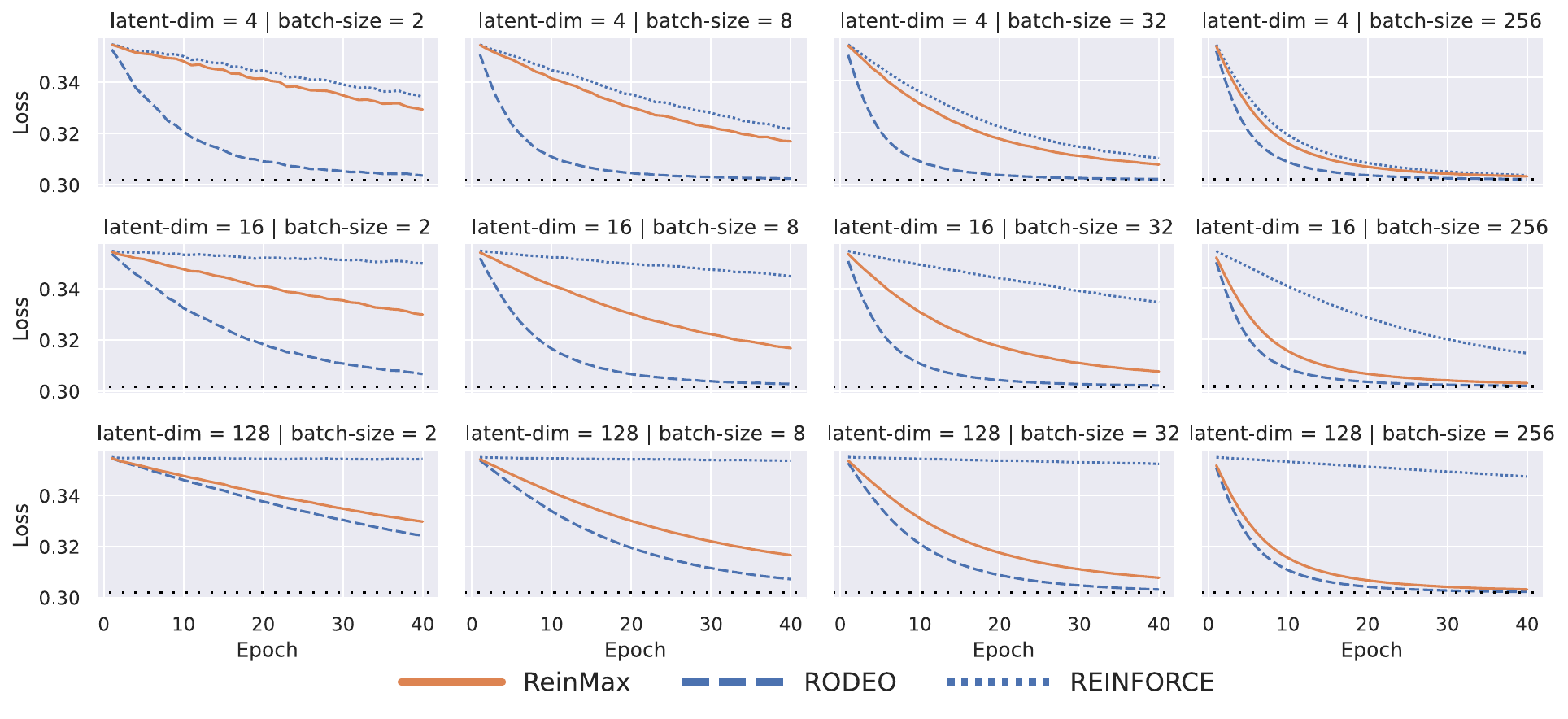}
        \vspace{-4mm}
        \caption{$p=1.5$ and $\vc = \rrrl0.45, \cdots, 0.45\rrrr$.}
        \label{}
    \end{subfigure}
    \caption{
     Training curves of polynomial programming, i.e., $\min_\vtheta E_{\mX}\rrrl\frac{\|\mX - \vc\|_p^p}{L}\rrrr, \mbox{ where } \mX \in \{0, 1\}^{L}, \mX_i \overset{\mathrm{iid}}{\sim} \mbox{Multinomial}(\mbox{softmax}(\vtheta_i)), \vtheta = [\theta_1, \cdots, \theta_L]^T, \vtheta_i \in \mathbb{R}^{2}$, and $L$ is the number of latent dimensions. 
    }
    \label{fig:poly-simple-rebuttal}
\end{figure}

\subsection{Additional Results on MNIST-VAE}
In our discussions in Section~\ref{sect:experiment}, we focused on the training ELBO only. Here, we provide a brief discussion on the test ELBO. 

\smallsection{Choosing Hyper-parameter Based on Training Performance}
Similar to Table~\ref{table:mnist-vae}, for each method, we select the hyper-parameter based on its training performance. 
The Test $-$ELBO in this setting is summarized in Table~\ref{table:mnist-vae-test-overfitting}. 
Despite the model being trained without dropout or other overfitting reduction techniques, ReinMax maintained the best performance in this setting. 

\smallsection{Choosing Hyper-parameter Based on Test Performance}
We also conduct experiments by selecting hyper-parameters directly based on their test performance. 
In this setting, the test $-$ELBO is summarized in Table~\ref{table:mnist-vae-test-best}, and the training $-$ELBO is summarized in Table~\ref{table:mnist-vae-train-overfitting}. 
ReinMax achieves the best performance in all settings except the test performance of the setting with 10 categorical dimensions and 30 latent dimensions. 

\subsection{More Comparisons with RODEO}
\label{appendix:rodeo}
To better understand the difference between RODEO and ReinMax, we conduct more experiments on polynomial programming, i.e., $\min_\vtheta E_{X} \Large[\normalsize \frac{|\mX - \vc|_p^p}{L}\Large]\normalsize$. Specifically, we consider polynomial programming under two different settings that define $\vc$ differently:
\begin{itemize}[leftmargin=*]
    \item
    \vspace{-0.2cm}
    In setting A, we have $\vc = [0.45, \cdots, 0.45]$. This is the setting we used in the submission.
    
    \item 
    In setting B, we have $\vc = [\frac{0.5}{L}, \frac{1.5}{L}, \cdots, \frac{L-0.5}{L}]$.
\end{itemize}

As to the difference between the Setting A and the Setting B, we would like to note:
\begin{itemize}[leftmargin=*]
    \item
    \vspace{-0.2cm}
In setting A, since $\forall i, \vc_i=0.45$ and $\vtheta_i\sim \mbox{Uniform}(-0.01, 0.01)$ at initialization, $E_{\mX_i \sim \mbox{softmax}(\theta_i)}\Large[\normalsize \frac{|\mX_i - \vc_i|_p^p}{L}\Large]\normalsize$ would have similar values. Therefore, the optimal control variates for $\vtheta_i$ are similar across different $i$.

    \item 
In setting B, we set $\vc_i$ to different values for different $i$, and thus the optimal control variate for $\vtheta_i$ are different across different $i$.
Therefore, Setting A is a simpler setting for applying control variate to REINFORCE.
\end{itemize}
As in Figure~\ref{fig:poly-simple-rebuttal}, ReinMax achieves better performance in more challenging scenarios, i.e., smaller batch size, more latent variables, or more complicated problems (Setting B or VAEs). 
Meanwhile, REINFORCE and RODEO achieve better performance on simpler problem settings, i.e., larger batch size, fewer latent variables, or simpler problems (Setting A).

\end{document}